\pdfoutput=1

\documentclass[11pt]{article}

\usepackage[]{acl}

\usepackage{times}
\usepackage{latexsym}
\usepackage{changepage}
\usepackage{multirow}
\usepackage{booktabs}
\usepackage{xspace}
\usepackage{amsfonts}
\usepackage{enumitem}
\usepackage{amsmath}
\usepackage{amssymb}
\usepackage{microtype}
\usepackage{pifont}
\usepackage{tikz}
\usepackage{colortbl}
\usepackage{url}
\usepackage{dsfont}
\usepackage[textsize=scriptsize]{todonotes}

\usepackage{lipsum}
\newcommand\blfootnote[1]{%
  \begingroup
  \renewcommand\thefootnote{}\footnote{#1}%
  \addtocounter{footnote}{-1}%
  \endgroup
}

\definecolor{red1}{HTML}{C82506}
\definecolor{blue1}{HTML}{7E91D3}
\definecolor{orange1}{HTML}{FD8662}
\definecolor{red2}{HTML}{FFB6B8}
\definecolor{red3}{HTML}{C82506}
\definecolor{green2}{HTML}{BFD8B6}
\definecolor{green3}{HTML}{E7F0E5}
\definecolor{green4}{HTML}{579D42}
\definecolor{brown2}{HTML}{DFC7B7}
\definecolor{blue1}{HTML}{0365C0}

\newcommand{\cmark}{\ding{51}}
\newcommand{\xmark}{\ding{55}}
\newcommand{\notcheckmark}{{\cmark}\textsuperscript{\textcolor{black}{\kern-0.7em{\bf---}}}}

\newcommand{\logo}{\includegraphics[scale=0.018]{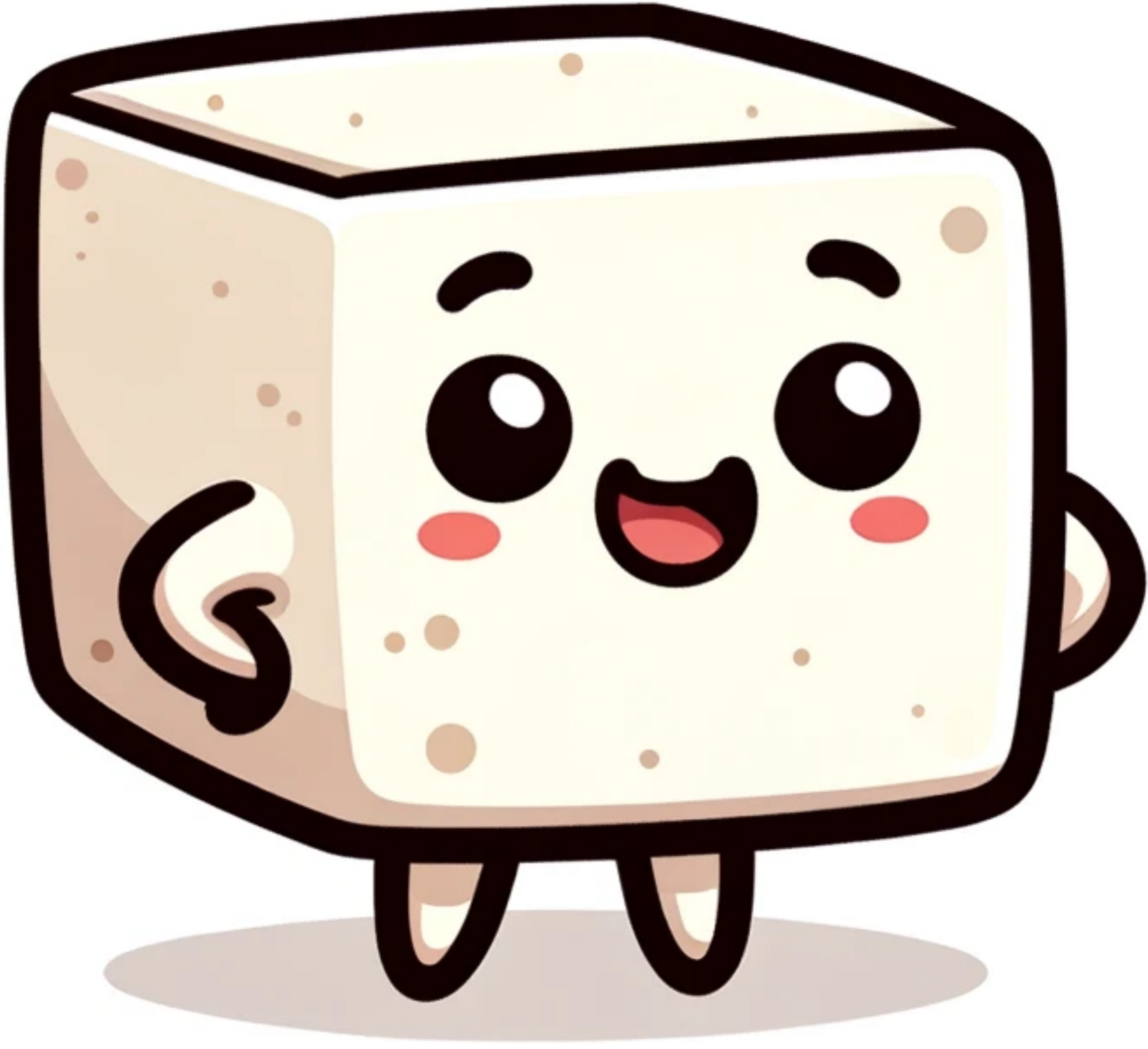}}

\usepackage[T1]{fontenc}

\usepackage[utf8]{inputenc}

\usepackage{microtype}

\title{\raisebox{-0.55ex}\logo\,\textsc{TofuEval}: Evaluating Hallucinations of LLMs \\ on Topic-Focused Dialogue Summarization}

\author{Liyan Tang$^{\diamondsuit \dagger}$, Igor Shalyminov$^\spadesuit$, Amy Wing-mei Wong$^\spadesuit$, Jon Burnsky$^\spadesuit$, Jake W. Vincent$^\spadesuit$ \\ \bf Yu'an Yang$^\spadesuit$, Siffi Singh$^\spadesuit$, Song Feng$^\spadesuit$, Hwanjun Song$^{\heartsuit \ddagger}$, Hang Su$^\spadesuit$, Lijia Sun$^\spadesuit$, \\ \bf Yi Zhang$^\spadesuit$, Saab Mansour$^\spadesuit$, Kathleen McKeown$^{\spadesuit}$ \\  $^\spadesuit$AWS AI Labs \quad $^{\heartsuit}$Korea Advanced Institute of Science \& Technology \\ $^{\diamondsuit}$The University of Texas at Austin \\
\texttt{shalymin@amazon.com}
}

\begin{document}
\maketitle

\begin{abstract}
Single document news summarization has seen substantial progress on faithfulness in recent years, driven by research on the evaluation of factual consistency, or hallucinations. We ask whether these advances carry over to other text summarization domains. We propose a new evaluation benchmark on topic-focused dialogue summarization, generated by LLMs of varying sizes. We provide binary sentence-level human annotations of the factual consistency of these summaries along with detailed explanations of factually inconsistent sentences. Our analysis shows that existing LLMs hallucinate significant amounts of factual errors in the dialogue domain, regardless of the model's size. On the other hand, when LLMs, including GPT-4, serve as binary factual evaluators, they perform poorly and can be outperformed by prevailing state-of-the-art specialized factuality evaluation metrics. Finally, we conducted an analysis of hallucination types with a curated error taxonomy. We find that there are diverse errors and error distributions in model-generated summaries and that non-LLM based metrics can capture all error types better than LLM-based evaluators.\footnote{We release the benchmark dataset with expert annotations at \href{https://github.com/amazon-science/tofueval}{\texttt{github.com/amazon-science/tofueval}}.}\blfootnote{$^\dagger$Work done as an intern at Amazon.}\blfootnote{$^\ddagger$Work done while at Amazon.}

\end{abstract}

\section{Introduction}

Recently, the field of automated text summarization has been increasingly inclined toward using Large Language Models (LLMs) to evaluate model outputs \cite{fu2023gptscore, gao-etal-2023-rarr, madaan2023selfrefine}. Given the consequential nature of this trend, we ask: \textbf{are LLMs up to the task of evaluating model outputs?} While recent work on news summarization has shown that LLMs' performance at evaluating the factual consistency of generated news summaries is promising \cite{luo2023chatgpt, wang2023chatgpt}, they may not perform as well in other less-explored summarization domains.

\begin{figure}
    \centering
\includegraphics[width=\linewidth, trim=0mm 0mm 00mm 0mm,clip]{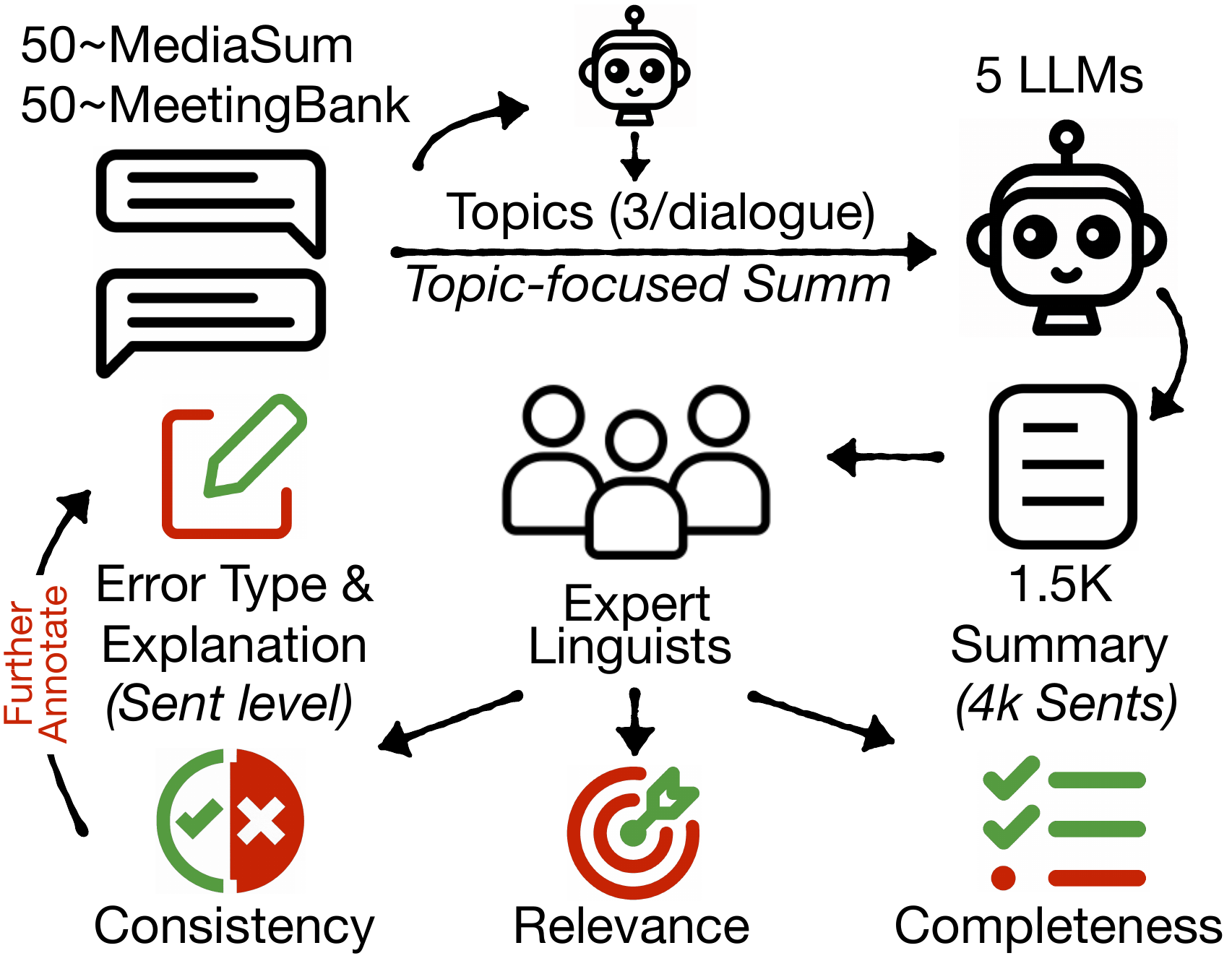}
    \caption{\textsc{TofuEval} contains 1.5K topic-focused summaries from two dialogue summarization datasets. We ask expert linguistic annotators to evaluate completeness, relevance and factual consistency of each summary, along with explanations and error types for factually inconsistent sentences.}
    \label{fig:main}
\end{figure}

Existing studies primarily investigate news summarization benchmarks, such as \citet{tang-etal-2023-understanding, laban-etal-2022-summac, pagnoni-etal-2021-understanding, fabbri-etal-2021-summeval}.
Alongside the finding that LLMs are capable of generating summaries of news articles that align with human preferences \cite{goyal2022news, Zhang2023BenchmarkingLL}, we ask: \textbf{can LLMs generate factually consistent summaries without hallucinations for non-news domains?} Given the potential benefits that dialogue summarization could bring to other areas, such as enhancing productivity in meetings or streamlining customer service interactions, we focus on dialogue summarization as a case study in this work.

We address the two questions mentioned above by introducing a new benchmark dataset \textbf{\textsc{TofuEval}}, which targets \textbf{To}pic-\textbf{f}oc\textbf{u}sed Dialogue summarization \textbf{Eval}uation of factual consistency. The benchmark dataset contains summaries generated by five LLMs of various sizes. Summaries in the benchmark are focused on specific topics in the dialogues due to the length of the dialogues and the fact that topics can hold varying degrees of importance to different users. 

In \textsc{TofuEval}, we engage professional linguistic data annotators to perform binary factuality evaluation of the sentences within each summary and write explanations for the sentences they determine to contain hallucinated contents (Section~\ref{sec:human-eval}). Human annotations reveal that LLMs are prone to making a substantial number of factual errors, and in contrast to common belief, larger LLMs do not necessarily generate more factually consistent dialogue summaries than smaller models (Section~\ref{sec:llm-summarizer}). 

Moreover, all LLMs we studied (including GPT-4), when used as binary factual consistency evaluators, perform poorly at detecting errors in LLM-generated summaries that focus on the main topic of a document according to human judgment (Section~\ref{sec:llm-as-evaluators}). In contrast, non-LLM-based factuality metrics demonstrate superior performance compared to most LLMs we tested, and they have the added advantage of smaller model size and lower inference costs. Our error analysis further reveals that those non-LLM metrics are better at capturing all error types when compared to LLMs.

\begin{table*}
\small
\centering
\renewcommand{\tabcolsep}{0.95mm}
\begin{tabular}{lcccccc|c}
\toprule
 & \textsc{SummEval} & \textsc{Frank} & \textsc{SummaC} & \textsc{AggreFact} & \textsc{DialEval} & \textsc{SummEdits} & \textbf{\textsc{TofuEval}} \\
Summaries from LLMs & \cellcolor{red2}\xmark & \cellcolor{red2}\xmark & \cellcolor{red2}\xmark & \cellcolor{red2}\xmark & \cellcolor{red2}\xmark & \cellcolor{green2}\cmark & \cellcolor{green2}\cmark \\
Non-Edited Summaries & \cellcolor{green2}\cmark & \cellcolor{green2}\cmark & \cellcolor{green2}\cmark & \cellcolor{green2}\cmark & \cellcolor{green2}\cmark & \cellcolor{red2}\xmark & \cellcolor{green2}\cmark \\
Fine-Grained Annotations & \cellcolor{red2}\xmark & \cellcolor{green2}\cmark & \cellcolor{green2}\notcheckmark & \cellcolor{green2}\notcheckmark & \cellcolor{red2}\xmark & \cellcolor{red2}\xmark & \cellcolor{green2}\cmark \\
Natural Language Explanations & \cellcolor{red2}\xmark & \cellcolor{red2}\xmark & \cellcolor{red2}\xmark & \cellcolor{red2}\xmark & \cellcolor{red2}\xmark & \cellcolor{red2}\xmark & \cellcolor{green2}\cmark \\
Document Domain & news & news & news & news & dialogue & mixed & dialogue \\
Summary Type & generic & generic & generic & generic & generic & generic & topic-focused \\
Annotators & crowd & crowd & mixed & mixed & students & trained ann. & linguists \\
Average Document Length & 408 & 595 & 583 & 496 & 130 & 705 & 950 \\ 
\bottomrule
\end{tabular}
\caption{\textbf{Comparison between \textsc{TofuEval} and existing factual consistency evaluation benchmarks on text summarization.} Ours is the first that focuses on different topics within a document and provides expert-annotated factual consistency labels for summary sentences with written explanations. We consider sentence-level and more granular annotations as fine-grained annotations. Some datasets in \textsc{SummaC} and \textsc{AggreFact} include this type of annotation partially (\notcheckmark). \textsc{DialEval} stands for \textsc{DialSummEval}.} \label{sec:benchmark-compare}
\end{table*}

Our contributions can be summarized as follows: (1) we are the first to introduce a topic-focused dialogue summarization benchmark \textsc{TofuEval} for factual consistency evaluation, which consists of LLM-generated summaries with \emph{expert-annotated factuality labels and explanations}; (2) we systematically evaluate LLMs as summarizers across relevance, completeness, and factual consistency, and we show that LLMs perform poorly on factual consistency in the dialogue domain; (3) on factuality prediction, our evaluation benchmark shows that with the exception of GPT-4, all other LLM-based evaluator performances we studied are inferior to non-LLM factuality metrics; (4) we conduct an error analysis using a curated error taxonomy, revealing that non-LLM factuality metrics can capture all error types better than LLM-based evaluators;
(5) we release \textsc{TofuEval} with human-annotated data to enable further research into improved automated evaluation of summary factuality.

\section{Related Work}

\paragraph{Factual Consistency Evaluation Benchmarks} In text summarization, there have been significant efforts to collect human-annotated data for assessing the effectiveness and correlation of different evaluation metrics with human judgments in detecting hallucianted contents in generated summaries \cite{fabbri-etal-2021-summeval, cao-wang-2021-cliff, maynez-etal-2020-faithfulness}.\footnote{We use the terms \emph{factual inconsistency}, \emph{factual errors} and \emph{hallucinations} interchangeably in this work.} 

Our proposed benchmark \textsc{TofuEval} aligns with these efforts but differs from prior work as follows (summarized in Table~\ref{sec:benchmark-compare}): (1) unlike existing evaluation benchmarks that contains non-LLM-generated summaries, \textsc{TofuEval} focuses on LLM-generated summaries. Contrasting with \textsc{SummEdits} \cite{laban-etal-2023-summedits}, which produces factually inconsistent summaries by editing correct LLM outputs, we directly identify factual errors in LLM-generated summaries. (2) \textsc{TofuEval} focuses on dialogue summarization. Even though \textsc{DialSummEval} \cite{gao-wan-2022-dialsummeval} shares this focus, source documents in \textsc{TofuEval} are considerably longer than those in \textsc{DialSummEval}, which are based on short conversations in the SAMSum corpus \cite{gliwa-etal-2019-samsum}. (3) Human evaluation from prior work comes from diverse sources, such as crowd-workers in \textsc{SummEval} \cite{fabbri-etal-2021-summeval} and \textsc{Frank} \cite{pagnoni-etal-2021-understanding}, and trained college students from \textsc{DialSummEval} \cite{gao-wan-2022-dialsummeval}. \textsc{TofuEval} consists of annotations from professional linguistic data annotators. 

\paragraph{Detecting Hallucinations}

Common automatic metrics for text summarization such as ROUGE \cite{lin-2004-rouge}, BLEU \cite{papineni-etal-2002-bleu}, and BERTScore \cite{bert-score} have poor correlations with human judgement on factual consistency \cite{kryscinski-etal-2019-neural, falke-etal-2019-ranking, gao-wan-2022-dialsummeval, Tang-etal-2023-medevidence}. Therefore, a few non-LLM-based metrics have been developed to detect factuality errors \cite{kryscinski-etal-2020-evaluating, goyal-durrett-2021-annotating, laban-etal-2022-summac, fabbri-etal-2022-qafacteval, zha-etal-2023-alignscore}. More recently, LLMs have been shown to have superior zero-shot performance at factual consistency evaluation under certain evaluation settings, highlighting their potential as state-of-the-art factual consistency evaluators \cite{luo2023chatgpt, wang2023chatgpt}.

As hallucinations from more recent models are harder to detect \cite{tang-etal-2023-understanding}, we re-evaluate non-LLM-based and LLM-based factuality metrics using LLM-generated summaries within the context of our dialogue summarization benchmark \textsc{TofuEval}. We find that non-LLM-based metrics can surpass most LLM-based evaluators. Nevertheless, all automated factuality metrics still perform quite poorly, underlining the challenging nature of the problem and the substantial room for improvement in automated factual inconsistency detection.

\section{\textsc{TofuEval} Benchmark} \label{sec:benchmark-creation}

Our topic-focused dialogue summarization benchmark \textsc{TofuEval} is constructed as follows: (1) sample documents from two publicly available dialogue summarization datasets (Section~\ref{sec:doc-select}); (2) create 
a variety of topics for sampled documents (Section~\ref{sec:topic-gen}); and (3) generate topic-focused summaries with various LLMs (Section~\ref{sec:summ-gen}). The resulting benchmark contains 100 dialogues and 15 LLM-generated summaries per dialogue; (4) lastly, we provide fine-grained human annotations on the topics and the generated summaries for dimensions including factual consistency, relevance, and completeness (Section~\ref{sec:human-eval}). The dataset construction pipeline is illustrated in Figure~\ref{fig:main}. 

\subsection{Document Selection} \label{sec:doc-select}

We select documents from two publicly available dialogue summarization datasets:

\paragraph{MediaSum} \cite{zhu-etal-2021-mediasum} is a large-scale dialogue summarization dataset with public interview transcripts from NPR and CNN. The dataset features multi-party conversations across various subjects, such as politics, economics, and US news.

\paragraph{MeetingBank} \cite{hu-etal-2023-meetingbank} is a summarization dataset with city council meetings. These meetings involve discussion and decisions about a diverse range subjects central to local governance and community welfare, including budget allocation, infrastructure planning, and crime prevention.

In our sampling process, we opt for documents with lengths ranging from 800 to 1,200 words. This decision was made to ensure that (1) the selected document lengths fit the maximum input size of all the models being evaluated and (2) the documents were sufficiently long to potentially elicit factual inconsistency errors in LLM-generated summaries. Opting for longer documents might pose challenges on manual evaluation. The benchmark statistics are shown in Table~\ref{sec:benchmark-stats}. We randomly sample 50 documents from the original test splits of each of these datasets for the benchmark construction.

\subsection{Topic Generation} \label{sec:topic-gen}

The impressive performance of LLMs enables the generation of a variety of summaries for a single long dialogue based on different points of interest in the dialogue with varying degrees of importance to readers. Instead of asking for generic summaries (\emph{i.e.}, summarizing a document in a few sentences), the performance of which has already been heavily evaluated (Table~\ref{sec:benchmark-compare}), we evaluate LLMs' performance in generating factually consistent summaries for specific topics within the sampled documents. Here we broadly define a \emph{topic} as a subject of discussion in a document related to an entity, an event, or an issue that readers of the document would like to know about \cite{Halliday2014}.

Since MediaSum and MeetingBank do not come with human-written topics, and identifying topics manually is time-consuming, we chose to identify main topics in a docuement with an LLM using a zero-shot prompt in Appendix~\ref{sec:prompt-topic}. We generated three topics for each document.\footnote{Given the length of the dialogues in \textsc{TofuEval}, we restrict the number of topics to three for each document.} Note that although we prompt the LLM to generate main topics, our human evaluation (more details in Section~\ref{sec:human-eval}) shows that  while the majority of LLM-generated topics are closely relevant, a small proportion of our collected topics are marginally within the context of the document. We decided to retain these marginal topics based on the assumption that marginal topics can also be useful to summary readers.

\subsection{Summarization Model Selection} \label{sec:summ-gen}

We construct the summarization factual consistency evaluation benchmark based on summaries generated by LLMs. This enables the creation of multiple summaries per dialogue and thus allows for easy scaling of the dataset with less human effort.

We chose to evaluate the summarization performance of one proprietary LLM, OpenAI's \textbf{GPT-3.5-Turbo}, and four open-source models, \textbf{Vicuna-7B} \cite{vicuna2023} and \textbf{WizardLM-7B/13B/30B} \cite{xu2023wizardlm}. More details about the models and our model selection can be found in Appendix~\ref{sec:model-selection}. We used a zero-shot prompt in Appendix~\ref{sec:tofu-prompt} for topic-focused text summarization. Unless otherwise stated, we set the model temperature to 0.7 and left the values of the remaining hyper-parameters unchanged for all models across all experiment settings in this work.

\paragraph{Dataset Splits} In summary, \textsc{TofuEval} consists of 50 documents per dataset, 3 generated topics per document, and 5 summaries from diverse LLMs per topic, resulting in 50$\times$2$\times$3$\times$5 = 1,500 summaries (refer to Table~\ref{sec:benchmark-stats} for more details). Further, we removed 23 model outputs that were deemed as non-summaries by human annotators, resulting in 1,479 summaries (3,966 summary sentences). We then randomly split the benchmark into development and test sets, with a 70\%/30\% development/test partition split on distinct documents.

\subsection{Annotation Collection} \label{sec:human-eval}

Using generated summaries, we collected high-quality annotations from professional linguistic data annotators for each dimension defined below.\footnote{We do not evaluate fluency and coherence since LLMs generally excel in these dimensions \cite{goyal2022news, Zhang2023BenchmarkingLL}.} The full annotation instructions and details about our quality control can be found in Appendix~\ref{sec:annotation-instruction}. 

\paragraph{Topic Categorization} 

We manually categorized topics within a document into \emph{main} and \emph{marginal} topics. Main topics refer to central information that is being discussed or presented in the document. Marginal topics are those that are explored less thoroughly in the documents. More detailed definitions can be found in Appendix~\ref{sec:Annotation-task-1}. Main topics make up approximately 75\% of the topics in \textsc{TofuEval} according to our categorization results (Table~\ref{sec:benchmark-stats}).

\paragraph{Factual Consistency} A summary sentence is factually consistent with a document if the sentence is stated or implied by the document; otherwise, it is inconsistent. For any sentences deemed inconsistent, the annotator wrote a brief explanation about the inconsistency. We aggregate sentence-level binary factuality labels to obtain labels for the entire corresponding summary: a summary is factually consistent with the document if all summary sentences are labeled as factually consistent; otherwise, the summary is factually inconsistent. 

\paragraph{Relevance} A relevant summary focuses on topic-related content from a source document. Each summary was assigned a relevance score ranging from 0 to 1, with 1 indicating an on-topic summary.

\paragraph{Completeness} A complete summary summarizes all information in the document that is related to the topic. Each summary was assigned a completeness score ranging from 0 to 1, with 1 indicating the highest level of completeness (Appendix~\ref{sec:Annotation-task-2}).

\subsection{Dialogue Summarization vs News Summarization}

Compared to news summarization, dialogue summarization involves unique challenges due to the informal and colloquial nature of dialogues, which requires summarization models to handle subtleties and noises. Additionally, dialogues are inherently interactive, which often involves a mix of questions, answers, and opinions among different speakers. This interaction requires a sophisticated understanding by the models of the contextual relationships between the pieces of information discussed in the dialogue. These complexities make dialogue summarization challenging and susceptible to factual inconsistencies (Section~\ref{sec:llm-summarizer}). This further makes it difficult to identify hallucinations in generated summaries in \textsc{TofuEval} (Section~\ref{sec:llm-as-evaluators}).

\section{Results: LLMs as Summarizers} \label{sec:llm-summarizer}

\begin{table*}
\small
\centering
\begin{tabular}{lcccc|cccc}
\toprule
\multirow{3}{*}{\textbf{\begin{tabular}[c]{@{}l@{}}Summ.\\ Model\end{tabular}}} & \multicolumn{4}{c}{\textbf{Sentence-Level (\% Error)}} & \multicolumn{4}{c}{\textbf{Summary-Level (\% Error)}} \\
\cmidrule(r){2-5} \cmidrule(r){6-9} 
 & \multicolumn{2}{c}{\textbf{MediaSum}} & \multicolumn{2}{c}{\textbf{Meetingbank}} & \multicolumn{2}{c}{\textbf{MediaSum}} & \multicolumn{2}{c}{\textbf{Meetingbank}} \\
\cmidrule(r){2-3} \cmidrule(r){4-5} \cmidrule(r){6-7} \cmidrule(r){8-9}
 & \textbf{Main} & \textbf{Marginal} & \textbf{Main} & \textbf{Marginal} & \textbf{Main} & \textbf{Marginal} & \textbf{Main} & \textbf{Marginal} \\
\midrule
Vicuna-7B & 19.6 & 35.8 & 17.6 & 36.8 & 42.7 & 55.4 & 33.0 & 58.0 \\
WizardLM-7B & 29.1 & 36.4 & 21.3 & 42.4 & 49.6 & 54.8 & 35.6 & 49.0 \\
WizardLM-13B & 17.4 & \cellcolor{green3} 27.2 & 15.8 & \cellcolor{green3} 25.4 & \cellcolor{green3}35.9 & \cellcolor{green3}44.4 & 41.3 & 46.8 \\
WizardLM-30B & \cellcolor{green3}14.6 & \cellcolor{green3} 27.2 & \cellcolor{green3} 13.7 & 26.2 & \cellcolor{green3}35.9 & 48.2 & \cellcolor{green3} 31.5 & \cellcolor{green3} 44.8 \\
GPT-3.5-Turbo & \cellcolor{green2}8.8 & \cellcolor{green2}13.6 & \cellcolor{green2}4.4 & \cellcolor{green2}9.4 & \cellcolor{green2}22.2 & \cellcolor{green2}27.2 & \cellcolor{green2}10.9 & \cellcolor{green2}19.8 \\
\midrule
Average & 17.5 & 27.8 & 14.4 & 27.8 & 37.2 & 46.0 & 30.4 & 43.6 \\
\bottomrule
\end{tabular}
\caption{\textbf{Percentage of sentence/summary-level factual inconsistencies across the five models used in \textsc{TofuEval}.} We show the error rates for main-topic summaries separately from those for marginal-topic summaries. We highlight the \colorbox{green2}{lowest} and \colorbox{green3}{second lowest} error rates. See examples of annotated summaries in Table~\ref{tab:example-1} and~\ref{tab:example-2}.} \label{tab:topic_summ_stats}
\end{table*}

\begin{figure*}
    \centering
\includegraphics[width=\linewidth, trim=0mm 0mm 00mm 0mm,clip]{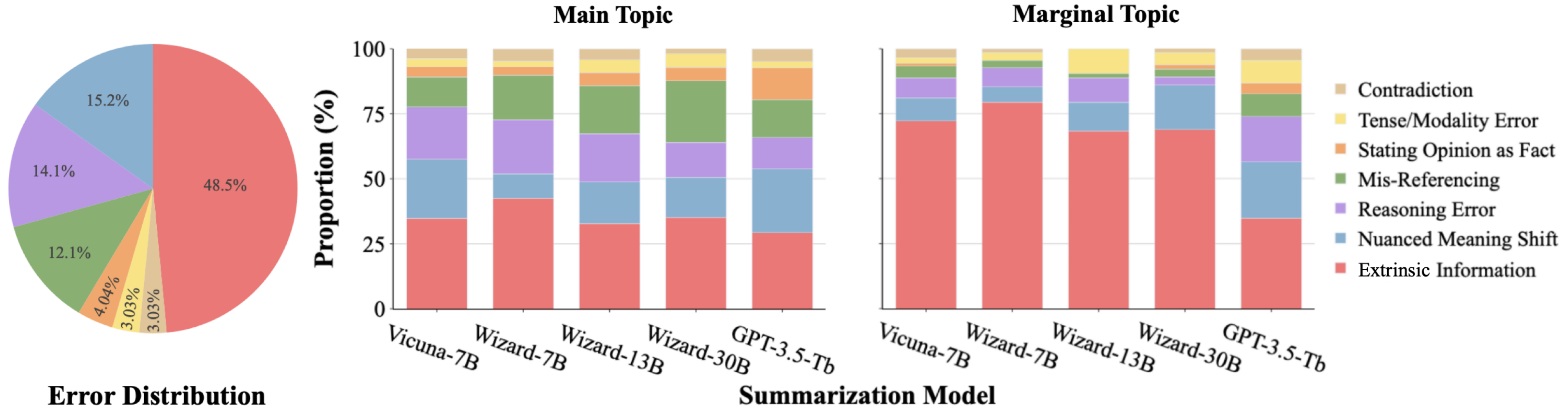}
    \caption{Error distribution \emph{over factually inconsistent summary sentences} for \textsc{TofuEval} (left) and for each summarizer over main/marginal topics (right). See error distributions \emph{over all summary sentences} for each summarizer over main/marginal topics in Appendix Figure~\ref{fig:error_dist_plot_appendix}.}
    \label{fig:error_dist_plot}
\end{figure*}

We show the error rate in generated summaries in Table~\ref{tab:topic_summ_stats} on both main and marginal topics. \textbf{Overall, LLMs we studied make a significant amount of factual errors, especially on the summary level.}

\begin{figure*}
    \centering
\includegraphics[width=\linewidth, trim=0mm 0mm 00mm 0mm,clip]{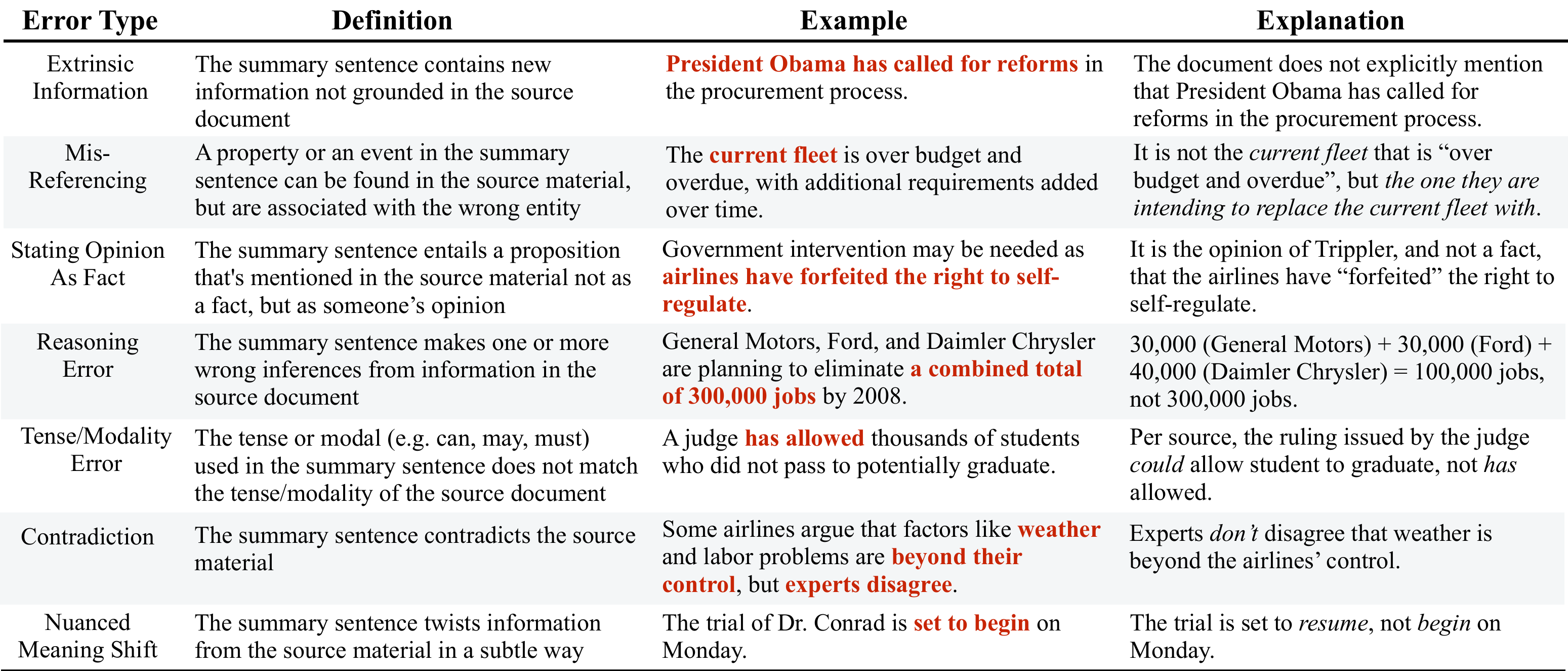}
    \caption{\textbf{Error taxonomy and definitions.} We include examples of factually inconsistent summary sentences and corresponding human annotated explanations from \textsc{TofuEval}. {\color{red1}{\textbf{Error spans}}} are highlighted (not included in \textsc{TofuEval}).}
    \label{fig:error_type}
\end{figure*}

We further investigate the distribution of different hallucination types in \textsc{TofuEval} with our curated error taxonomy. Note that our taxonomy closely resembles that of \citet{tang-etal-2022-confit}, which is based on the SAMSum dialogue dataset \cite{gliwa-etal-2019-samsum}. Due to the complexity of the long dialogues in \textsc{TofuEval}, we extend the taxonomy of \citet{tang-etal-2022-confit} with new error types, such as  \emph{reasoning error} and \emph{stating opinion as fact}. A summary of our curated error taxonomy for the benchmark is provided in Figure~\ref{fig:error_type}. We leverage the error taxonomy to enrich all binary factual inconsistency annotations in the benchmark. Additional details about the taxonomy curation process and error-type annotation can be found in Appendix~\ref{sec:err-type-anno}.

\paragraph{LLMs tend to produce more factually inconsistent summaries when prompted to focus on a marginal topic, especially with extrinsic information error.} 
As shown in Figure~\ref{fig:error_dist_plot}, when prompting models for summaries about marginal topics, all summarizers generate significantly more \emph{Extrinsic Information}. We find that when the topic is barely mentioned in the document, models try to rely on their knowledge to make informed inferences about the topic, bringing unsupported information into the summary. An exception is GPT-3.5-Turbo, which generates far fewer Extrinsic Information for marginal topics. Compared to other summarizers that generate unsupported sentences, we find that GPT-3.5-Turbo often handles requests for marginal-topic summaries by including off-topic content or occasionally explicitly saying, ``\emph{the topic is not discussed in the document}''.\footnote{Further optimization of prompts to reduce the error rate for specific error types is beyond the scope of the current work.}

More findings can be found in Appendix~\ref{sec:extend-summarizer}.

\section{Results: LLMs as Evaluators} \label{sec:llm-as-evaluators}

We now move on to consider the use of LLMs as \emph{evaluators} of factual consistency rather than as summarizers. We first present an evaluation of their performance at making binary factual consistency predictions for both summaries and summary sentences (Section~\ref{sec:predict-label}). We then provide an error-type analysis, investigating which error types models fail to detect well (section~\ref{sec:metric-error-analysis}). Finally, given that LLMs have the ability to generate critiques of model output and provide explanations for their factual consistency judgments \cite{madaan2023selfrefine, saunders2022self}, we consider the accuracy of model-generated explanations by comparing them to human-written explanations (Appendix~\ref{sec:exp-eval}).

\paragraph{Evaluator Selection} For a comprehensive comparison, we include the following non-LLM based SOTA factuality metrics: 
SummaC-ZS, SummaC-CV \cite{laban-etal-2022-summac}, QAFactEval \cite{fabbri-etal-2022-qafacteval}, and AlignScore \cite{zha-etal-2023-alignscore}. We also include the following proprietary and open-source LLMs as factual consistency evaluators: (1) GPT-4 \cite{OpenAI2023GPT4TR}; (2) GPT-3.5-Turbo; (3) Vicuna-13B/33B \cite{vicuna2023}; (4) WizardLM-13B/30B \cite{xu2023wizardlm}. For all LLMs in the aforementioned list we used a zero-shot configuration, as certain LLMs in this list are unable to accommodate few-shot evaluations due to input length constraints. In any case, it has been observed that few-shot examples do not consistently yield superior outcomes in comparison to zero-shot scenarios \cite{laban-etal-2023-summedits, luo2023chatgpt}. More details about model selection are in Appendix~\ref{sec:model-selection}.

\subsection{Predicting Factual Consistency} \label{sec:predict-label}

We first measure the performance of the selected factual consistency evaluator models via a binary prediction task. For any evaluation model $M$, a dialogue $d$, and some generated content $c$, we ask $M$ to predict whether $c$ is factually consistent with the corresponding dialogue $d$:
$$M(d, c) \in \{\mathrm{consistent}, \mathrm{inconsistent}\}.$$ 
Following \citet{laban-etal-2022-summac,fabbri-etal-2022-qafacteval, tang-etal-2023-understanding, luo2023chatgpt}, we measured models' performance using the balanced accuracy (BAcc) method, which takes into account the imbalance of factually consistent and factually inconsistent summaries. We analyzed the results based on both sentence-level and summary-level prediction performance. Unless stated otherwise, all evaluation results shown here are based on the test set.

\paragraph{Obtaining Predictions from non-LLM based Factuality Metrics} The non-LLM-based models we used take as input a source and a piece of summary text to be evaluated, and they return a score for the evaluated text within a particular range of continuous values. Higher scores suggest more consistent summaries. Following \citet{laban-etal-2022-summac}, we decided on a threshold value for each metric using the development set and report the test set results assuming the selected threshold. We chose the thresholds for sentence-level and summary-level evaluations separately. Text that receives a value above the threshold is considered factually consistent with the document; otherwise, it is considered inconsistent. For our sentence-level and summary-level evaluations, the input text was a summary sentence and a whole summary, respectively.

\begin{table*}
\small
\centering
\begin{tabular}{clcccc|cccc}
\toprule
\multicolumn{1}{l}{\multirow{3}{*}{\textbf{\begin{tabular}[c]{@{}l@{}}Model\\ Type\end{tabular}}}} & \multirow{3}{*}{\textbf{\begin{tabular}[c]{@{}l@{}}Evaluation\\ Model\end{tabular}}} & \multicolumn{4}{c}{\textbf{Sentence-Level (BAcc $\uparrow$)}} & \multicolumn{4}{c}{\textbf{Summary-Level (BAcc $\uparrow$)}}  \\
\cmidrule(r){3-6} \cmidrule(r){7-10}
\multicolumn{1}{l}{} &  & \multicolumn{2}{c}{\textbf{MediaSum}} & \multicolumn{2}{c}{\textbf{MeetingBank}} & \multicolumn{2}{c}{\textbf{MediaSum}} & \multicolumn{2}{c}{\textbf{MeetingBank}} \\
\cmidrule(r){3-4} \cmidrule(r){5-6} \cmidrule(r){7-8} \cmidrule(r){9-10}
\multicolumn{1}{l}{} &  & \textbf{Main} & \textbf{Marginal} & \textbf{Main} & \textbf{Marginal} & \textbf{Main} & \textbf{Marginal} & \textbf{Main} & \textbf{Marginal} \\
\midrule
- & Baseline & 50.0 & 50.0 & 50.0 & 50.0 & 50.0 & 50.0 & 50.0 & 50.0  \\
\midrule
\multirow{4}{*}{\textbf{\begin{tabular}[c]{@{}c@{}}Non-\\ LLM\end{tabular}}} & SummaC-ZS & 66.1 & 73.9 & \cellcolor{green3}63.9 & \cellcolor{green3}80.6 & 62.7 & 64.1 & 58.1 & \cellcolor{green3} 72.4  \\
 & SummaC-CV & \cellcolor{green3}67.6 & 73.0 & 62.6 & 77.3 & 61.2 & 66.5 & 52.4 & \cellcolor{green3} 72.9  \\
 & QAFactEval & 53.9 & 74.0 & 58.0 & 75.8 & 61.4 & \cellcolor{green3}74.2 & 55.1 & 68.2  \\
 & AlignScore & \cellcolor{green2}69.2 & \cellcolor{green3}76.2 & 61.2 & 78.6 & \cellcolor{green2}65.5 & 72.1 & \cellcolor{green3}63.4 & 71.8 \\
\midrule
\multirow{4}{*}{\textbf{\begin{tabular}[c]{@{}c@{}}Open\\ Source\\ LLM\end{tabular}}} & Vicuna-13B & 54.0 & 54.8 & 49.6 & 61.9 & 55.6 & 59.1 & 51.2 & 59.2   \\
 & Vicuna-33B & 51.0 & 51.1& 53.6 & 48.4 & 52.5 & 53.4 & 53.2 & 51.0   \\
 & WizardLM-13B & 59.8 & 53.5 & 58.8 & 56.6 & 57.0 & 54.5 & 54.6 & 58.0   \\
 & WizardLM-30B & 54.5 & 53.9 & 53.5 & 53.4 & 53.3 & 54.4 & 53.0 & 53.2  \\
\midrule
\multirow{2}{*}{\textbf{\begin{tabular}[c]{@{}c@{}}Prop.\\ LLM\end{tabular}}} 
 & GPT-3.5-Turbo & 61.6 & 68.9 & 56.0 & 65.0 & 59.6 & 65.8 & \cellcolor{green3}63.2 & 65.7   \\
 & GPT-4 & 64.9 & \cellcolor{green2}80.2 & \cellcolor{green2}67.5 & \cellcolor{green2}90.3 & \cellcolor{green3}63.7 & \cellcolor{green2}78.9 & \cellcolor{green2}74.7 & \cellcolor{green2}83.1   \\
\bottomrule
\end{tabular}
\caption{\textbf{Sentence-level and summary-level balanced accuracy (BAcc) for factual consistency evaluators on the test set of \textsc{TofuEval}.} For LLM-based methods, we show summary-level labels by aggregating sentence-level labels, as it achieves better performance than directly predicting consistency labels on whole summaries. All results for LLMs are the average of three runs. Note that a \emph{baseline} method that always predicts inconsistent or consistent achieves 50\% balanced accuracy. We highlight the \colorbox{green2}{best} and \colorbox{green3}{second best} results. Prediction results for both \textbf{Main}-topic summaries and \textbf{Marginal}-topic summaries are shown.} \label{sec:eval-performance}
\end{table*}

\paragraph{Obtaining Predictions and Explanations from LLMs} We tested two methods for obtaining factual consistency labels. First, we directly asked LLMs to provide binary labels (\textsc{Dir}). In Table~\ref{sec:eval-performance}, we show the results obtained by a sentence-level prompt for all LLM-based metrics for both sentence-level and summary-level evaluation.\footnote{See Section~\ref{sec:human-eval} for how we obtained summary-level labels from sentence-level labels. Performance for the summary-level prompt can be found in Appendix Table~\ref{sec:summ-level-eval-performance}.} We run all LLMs \emph{three times} per completed prompt and report the average performance.

Next, to obtain explanations from LLMs, we attempted to elicit Chain-of-Thought reasoning following \cite{wei2022cot}. We adjusted the previous prompt, asking the LLM to provide explanations for its decisions in addition to providing binary judgments (\textsc{Exp}). We extracted binary predictions from the outputs of this prompt for model self-agreement evaluation (presented later in this section), and we extracted explanations for explanation evaluation (Appendix~\ref{sec:exp-eval}).\footnote{Binary prediction performance of this prompt can be found in Appendix Table~\ref{sec:summ-exp-level-eval-performance}, where we show that prompting for explanations does not improve performance on the binary prediction task.} Prompts for summary-level and sentence-level evaluation for both methods can be found in Appendix~\ref{sec:prompt-summ}.

\paragraph{Non-LLM factual consistency evaluation models perform well.} 
As shown in Table~\ref{sec:eval-performance}, GPT-4 achieves the best performance when it comes to factual consistency evaluation across datasets and topic types most of the time. Further, most of the second-best evaluators are the non-LLM-based factuality metrics across all configurations, outperforming LLMs, including GPT-3.5-Turbo, by a large margin. When evaluating the main-topic summaries of MediaSum data, AlignScore even surpasses GPT-4 in performance at both the sentence level and the summary level. It is worth noting that the non-LLM-based evaluators have faster inference speed, cost much less (compared to API calls), and only need a 16GB or smaller GPU to complete the prediction task.

For the open-source LLMs we tested, the balanced accuracy scores are all between 50\% and 60\%, which is barely better than the baseline. Although these models are shown to generate outputs preferred by humans compared to the proprietary models over a variety of instruction-following tasks \cite{xu2023wizardlm, alpaca_eval}, they are not equipped with the discrimination skills sufficient to perform this task well. Further, \textbf{larger open-source models do not outperform their smaller counterparts on most settings.} For example, Vicuna-33B's performance is 13\% worse than Vicuna-13B on marginal-topic summaries of MeetingBank data, and it is even worse than the baseline. Some possible explanations for this might include that these models are not pre-trained on this type of data, or they are not large enough to develop the emergent discrimination capabilities for this type of task compared to the proprietary models.

Overall, these findings \textbf{raise caution against unquestioning admiration for using cutting-edge LLMs as evaluators.}

\paragraph{It is more challenging for all models tested to detect errors in main-topic summaries.} As shown in Table~\ref{sec:eval-performance}, for both non-LLM-based factuality metrics and proprietary LLMs, they are on average about 10\% worse at detecting errors from main-topic summaries, whereas the best model, GPT-4, has a performance gap of approximately 10-20\% on the sentence-level prediction task for both datasets. We hypothesize that this is due to the fact that main-topic summaries do not contain a large proportion of \emph{extrinsic information} (Figure~\ref{fig:error_dist_plot}) which we find models can detect relatively well (Section~\ref{sec:metric-error-analysis}). As mentioned previously, the open-source LLMs we tested are not equipped with the skills necessary to perform this discrimination task, hence we do not notice any consistent performance improvement on the marginal-topic summaries, which seems slightly easier for other model types. \textbf{Overall, there is still a large room for improvement when it comes to efficient, cost-effective factual inconsistency detection}, especially for main-topic summaries, where existing models' performance is quite close to baseline performance which always predicts inconsistent or consistent. We explore differences in the error types that models fail to identify in Section \ref{sec:metric-error-analysis}.

\paragraph{Most LLMs, especially the smaller ones, lack consistency over multiple predictions generated with the same prompt.} We calculate each model's \emph{self-agreement} by comparing its predictions on the factuality of summary sentences.\footnote{Because proprietary LLMs do not provide token probabilities like open-source LLMs, for a fair comparison we chose to directly compare three runs from each model and calculate Cohen's kappa on the three predictions.} The sentence-level self agreement across the three runs for each model (as Cohen's kappa) is provided in Table~\ref{sec:self-agreement}, based on the binary prediction results from direct binary label predictions (\textsc{Dir}) and the label predictions with explanations (\textsc{Exp}) (Section~\ref{sec:predict-label}). We observe that GPT-4 has near-perfect agreement across all settings, suggesting that the model makes consistent predictions, while the remaining models have fair to moderate agreement ($\kappa$ between 0.2 and 0.6) most of the time for \textsc{Dir} predictions. Interestingly, we observe that asking the model for an explanation in addition to making a binary prediction on factuality lowers its Cohen's kappa score. This is more apparent for the smaller 13B models, where we have a maximum drop of 0.38 in agreement. We hypothesize that prompting models to generate explanations along with the binary prediction adds an extra layer of complexity to the task, yielding less deterministic results and causing lower self-agreement compared to the direct binary label prediction task.

\begin{table}
\small
\centering
\begin{tabular}{lcccc}
\toprule
\multirow{2}{*}{\textbf{\begin{tabular}[c]{@{}l@{}}Evaluation\\ Model\end{tabular}}} & \multicolumn{2}{c}{\textbf{MediaSum}} & \multicolumn{2}{c}{\textbf{Meetingbank}} \\
\cmidrule(r){2-3} \cmidrule(r){4-5}
 & \textbf{\textsc{Dir}} & \textbf{\textsc{Exp}} & \textbf{\textsc{Dir}} & \textbf{\textsc{Exp}} \\
\midrule
Vicuna-13B & 0.35 & 0.11 & 0.38 & 0.00 \\
Vicuna-33B & 0.37 & 0.18 & 0.29 & 0.13 \\
WizardLM-13B & 0.47 & 0.33 & 0.54 & 0.18 \\
WizadLM-33B & 0.50 & 0.39 & 0.35 & 0.34 \\
\midrule
GPT-3.5-Turbo & 0.57 & 0.44 & 0.59 & 0.51 \\
GPT-4 & \cellcolor{green2}0.96 & \cellcolor{green2}0.95 & \cellcolor{green2}0.90 & \cellcolor{green2}0.91 \\
\bottomrule
\end{tabular}
\caption{\textbf{Sentence-level model self-agreement in predicting factual consistency labels on the whole test set of \textsc{TofuEval}.} Models are run 3 times and self-agreements are calculated by Cohen's kappa.} \label{sec:self-agreement}
\end{table}

\begin{figure*}
    \centering
\includegraphics[width=\linewidth, trim=0mm 00mm 00mm 00mm,clip]{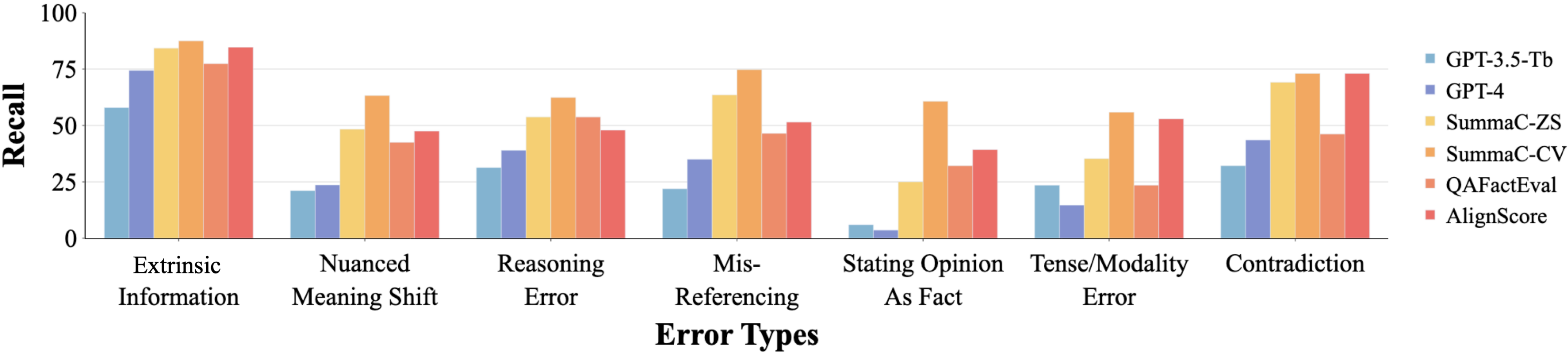}
    \caption{\textbf{Recall of summary factual inconsistency predictions by error types.} {\color{orange1}{\textbf{Non-LLM based factuality metrics}}} are better at capturing errors than {\color{blue1}{\textbf{LLM-based evaluators}}} across all error types.}
    \label{fig:model-recall}
\end{figure*}

\paragraph{There is a strong positive correlation between a model's self-agreement and its performance at factual consistency prediction.} We computed Pearson correlation coefficients $\rho$ to assess the relationship between models' performance (as BAcc) and their self agreement (as Cohen's Kappa) at the sentence level. The results revealed a substantial correlation on MediaSum data ($\rho$ = 0.79, $p$ = 0.02) and a highly significant correlation on MeetingBank data ($\rho$ = 0.99, $p$ = 0.00). In other words, there is a strong linear relationship between models' performance and their self agreement. When models exhibit greater self-consistency in predictions, they allocate a higher probability mass to these predictive labels. Given that this correlates well with the models' balanced accuracy, when thinking of these models altogether, we conclude that they are well-calibrated for this task.\footnote{Our insight here is based on configurations with a temperature of 0.7. This finding may not hold for other temperatures. For example, a temperature of zero would lead to deterministic outputs and hence perfect agreement, but balanced accuracy may still be low in such cases.}

More findings can be found in Appendix~\ref{sec:extend-llm-evaluator}.

\subsection{Error Analysis} \label{sec:metric-error-analysis}

Following \citet{tang-etal-2023-understanding}, we analyzed the evaluator models' error-type detection performance in terms of \emph{recall}.\footnote{Precision cannot technically be defined for each error type because evaluator models do not predict error types.} We divided all factually inconsistent summary sentences into several subsets, each of which contains only one error type. Given the evaluators' performance as shown in Table~\ref{sec:eval-performance}, we selected a subset of relatively strong evaluators for this analysis and show the result in Figure~\ref{fig:model-recall}.

\paragraph{Non-LLM based evaluation metrics are better at capturing all error types.} In Figure~\ref{fig:model-recall}, we show the performance of the LLM-based evaluators in blue and the non-LLM based evaluators in orange. We observe that all evaluators perform fairly well at identifying what we termed Extrinsic Information. This might be due to the fact that this type of error primarily involves unfamiliar noun phrases or events (relative to the document), which we suppose facilitates models' detection of this error type. That said, we do find that GPT-3.5-Turbo only detect 50\% of such errors---approximately 30\% lower than the rate for non-LLM based metrics. For the remaining error types, there is a large gap in the detection rate between LLM-based and non-LLM based metrics. It is possible that the tested LLMs may perform better at identifying certain error types with better prompt design, but we leave this for future work.

Note that \textbf{recall and balanced accuracy are complementary metrics that lend insight into evaluators' behavior} in our analysis. Having a high recall does not necessarily suggest high balanced accuracy, and vice versa. For example, although GPT-4 does not capture as many errors as non-LLM based models, it achieves a higher balanced accuracy (see Table~\ref{sec:eval-performance}). However, the non-LLM based metrics do surpass GPT-3.5-Turbo in both recall and balanced accuracy with most settings, suggesting that their performance is superior.

\section{Conclusion}

We have proposed a new factual consistency evaluation benchmark \textsc{TofuEval} for topic-focused dialogue summarization with LLM-generated summaries. We find that LLMs serving as summarizers make numerous and diverse hallucinations in their summaries. Furthermore, by measuring balanced accuracy and analyzing models' error types, we conclude that it is still challenging for both LLM-based evaluators and existing state-of-the-art non-LLM based factuality metrics to detect a wide range of  hallucinations in dialogue, although the latter exhibit slightly better performance.

\section*{Limitations}

Our work has a few limitations. First, in our proposed \textsc{TofuEval} benchmark, we do not ask human evaluators to annotate factual consistency errors that may span beyond a single sentence due to the already-high complexity of our annotation task. For example, one type of inter-sentential error is a discourse error, as discussed in \citet{pagnoni-etal-2021-understanding}. Secondly, our evaluation framework treats all factual errors as having equal severity, without distinguishing between the potentially-varying degrees of impact that different factual error types have. Thirdly, our summarization evaluation is specifically tailored for English dialogues. Models evaluated in this work may exhibit different performance for other domains and other languages. Additionally, we do not conduct extensive prompt engineering to identify better prompts for LLMs, which could lead to improvements in factual consistency or improved detection of factual errors. We leave this investigation to future work. Finally, we do not evaluate larger set of LLMs as factual consistency evaluators since GPT-3.5-Turbo and GPT-4 are shown to be representative about the recent LLMs' performance in factual consistency evaluation \cite{laban-etal-2023-summedits}. Despite the acknowledged limitations, we hope that our proposed benchmark and the insights will inspire future work focusing on enhancing automated evaluation metrics and the development of summarizers with better factual consistency performance.

\section*{Acknowledgements}

The authors wish to express our gratitude to our annotation team, whose vital contributions have been instrumental in this project: Marika Hall, Hoyeol Kim, Paul Goeden, Elvira Magomedova, Teddy Mutiga, Daniel North, Giuseppina Silverstri, Helen Satchwell, Anna Stallman, Aidan Thies, Michael Valentekovic, Jennifer Won, Carolina Kocuba, Neil Morrissey, Andy Bridges, Derek Chao, Mei Leung, Matthew Mahoney, Andrew McNally, Francis O'Brien, Alex Mowen, and Nicole LaManna. All the members of our annotation team are based in the U.S. and are paid a competitive hourly rate that is benchmarked against similar roles in the U.S.

\bibliography{anthology,custom}

\appendix

\section{\textsc{TofuEval} Descriptive Statistics}

We show the descriptive statistics of \textsc{TofuEval} in Table~\ref{sec:benchmark-stats}. We measure the word count by the \texttt{NLTK} package. All dialogues are written in English. 

\begin{table*}
\small
\centering
\renewcommand{\tabcolsep}{2mm}
\begin{tabular}{lccccccc}
\toprule
\textbf{Dataset} & \textbf{\begin{tabular}[c]{@{}c@{}}\#\\ Doc.\end{tabular}} & \textbf{\begin{tabular}[c]{@{}c@{}}Avg\\ Len\end{tabular}} & \textbf{\begin{tabular}[c]{@{}c@{}}\#Asp.\\ / Doc.\end{tabular}} & \textbf{\begin{tabular}[c]{@{}c@{}}\#\\ LLM\end{tabular}} & \textbf{\begin{tabular}[c]{@{}c@{}}Main\\ Topic\end{tabular}} & \textbf{\begin{tabular}[c]{@{}c@{}}\#\\ Turn\end{tabular}} & \textbf{\begin{tabular}[c]{@{}c@{}}\#\\ Speaker\end{tabular}} \\
\midrule
MediaSum & 50 & 970 & 3 & 5 & 78\% & 16.6 & 5.7\\
MeetingBank & 50 & 930 & 3 & 5 & 73\% & 19.9 & 4.9 \\
\bottomrule
\end{tabular}
\caption{\textbf{Dataset statistics on \textsc{TofuEval}.} We sample 50 test set documents from each dataset; generate 3 topics for each document and then collect summaries from 5 LLMs for each topic. We show the percentage of main topics evaluated by human (more in Section~\ref{sec:human-eval}).} \label{sec:benchmark-stats}
\end{table*}

\section{Model Details} \label{sec:model-selection}

\subsection{Summary Generation} 

We chose to use Vicuna-7B\footnote{\url{https://huggingface.co/lmsys/vicuna-7b-delta-v0}.}\cite{vicuna2023} and WizardLM-7B/13B/30B\footnote{\url{https://huggingface.co/WizardLM/WizardLM-7B-V1.0}, \url{https://huggingface.co/WizardLMTeam/WizardLM-13B-V1.0}, \url{https://huggingface.co/WizardLM/WizardLM-30B-V1.0}.} \cite{xu2023wizardlm}, both of which are based on Llama \cite{touvron2023llama}, for summary generation. We also experimented with other open-source LLMs, including Falcon-7b/40b-instruct\footnote{\url{https://huggingface.co/tiiuae/falcon-7b-instruct}, \url{ https://huggingface.co/tiiuae/falcon-40b-instruct}.} and mpt-7b-instruct\footnote{\url{mosaicml/mpt-7b-instruct}.}. We find that Vicuna and WizardLM generally do a better job at instruction following for our task and are more robust to prompts. We also collect summaries from GPT-3.5-Turbo via its official API.

Our model selection process for LLM-based summarization was finalized in early June for human evaluation, and as a result, we have not included summaries generated by models developed since then.

\subsection{Summary Evaluation}

In our study, we incorporate three SOTA and specialized factuality metrics based on entailment: SummaC-ZS, SummaC-CV\footnote{\url{https://github.com/tingofurro/summac/}} \cite{laban-etal-2022-summac}, and AlignScore\footnote{\url{https://github.com/yuh-zha/AlignScore}} \cite{zha-etal-2023-alignscore}. These metrics are designed to determine whether summary sentences can be inferred by some portion of text extracted from the source documents. We also include a SOTA question-answering (QA) based factuality metric QAFactEval\footnote{\url{https://github.com/salesforce/QAFactEval}} \cite{fabbri-etal-2022-qafacteval}, which evaluates factual consistency by generating questions and verifying the answerability of the generated questions. We refer readers to original works for more details of these models. For Vicuna-33B\footnote{\url{https://huggingface.co/lmsys/vicuna-33b-v1.3}}, we use the one based on Llama \cite{touvron2023llama}. We do not include instruction-tuned LLMs such as Falcon-40b and mpt-30b, as these models performed poorly in our initial benchmarks.

\section{Prompts} \label{sec:prompt-summ-eval}

\subsection{Prompt for Topic Generation} \label{sec:prompt-topic}

We use the following prompt to generate topics for dialogue documents in \textsc{TofuEval}.

\begin{adjustwidth}{1em}{1em}
\quad \emph{Document: \{Document\}}

\emph{Enumerate three main topics that people would like to know from the provided document. Each topic should be around 5 words.}
\end{adjustwidth}

\subsection{Prompt for Topic-Focused Summarization} \label{sec:tofu-prompt}

We add the following instruction to the model's default prefix, if any, to form the prompt\footnote{We also tried to control the summary length by asking models to generate a fixed number of sentences, but most models we evaluated here cannot follow the length constraint well in either format.} for topic-focused text summarization in a zero-shot manner:
\begin{adjustwidth}{1em}{1em}
\quad \emph{Document: \{Document\}}

\emph{Summarize the provided document focusing on ``\{topic\}''. The summary should be less than 50 words in length.}
\end{adjustwidth}

\subsection{Prompts for Summary Evaluation} \label{sec:prompt-summ}

This section contains all prompts that we used for obtaining binary factual consistency labels and explanations from LLMs. Additional details can be found in Section~\ref{sec:predict-label}. For sentence-level prompt, we find that \textbf{providing the previous summary sentences in the prompt as context does not affect the performance on an initial study}, so for simplicity, we only provided the isolated sentence.

\paragraph{(\textsc{Dir}) Binary-Label, Sentence-Level Prompt} We asked LLMs to provide a binary factual consistency label for a summary sentence using the following prompt:

\begin{adjustwidth}{1em}{1em}
\quad \emph{Document: \{Document\}}

\emph{Sentence: \{Sentence\}}

\emph{Determine if the sentence is factually consistent with the document provided above. A sentence is factually consistent with the document if it can be entailed (either stated or implied) by the document. Please answer with ``Yes'' or ``No''.}
\end{adjustwidth}

\paragraph{(\textsc{Dir}) Binary-Label, Summary-Level Prompt} We asked LLMs to provide a binary factual consistency label for a summary using the following prompt:
\begin{adjustwidth}{1em}{1em}
\quad \emph{Document: \{Document\}}

\emph{Summary: \{Summary\}}

\emph{Determine if the summary is factually consistent with the document provided above. A summary is factually consistent with the document if all information in the summary can be entailed (either stated or implied) by the document. Please answer with ``Yes'' or ``No''.}
\end{adjustwidth}

\paragraph{(\textsc{Exp}) Binary-Label with Explanation, Sentence-Level Prompt} We asked LLMs to provide explanations for their decisions in addition to providing the binary factuality judgment. Below is the prompt we used for sentence-level evaluation with corresponding explanations:
\begin{adjustwidth}{1em}{1em}
\quad \emph{Document: \{Document\}}

\emph{Sentence: \{Sentence\}}

\emph{Determine if the sentence is factually consistent with the document provided above. A sentence is factually consistent if it can be entailed (either stated or implied) by the document. Please start your answer with ``Yes.'' or ``No.'' Please briefly explain the reason within 50 words.}
\end{adjustwidth}

\paragraph{(\textsc{Exp}) Binary-Label with Explanation, Summary-Level Prompt} The following prompt was used for summary-level factuality evaluation with a corresponding explanation.
\begin{adjustwidth}{1em}{1em}
\quad \emph{Document: \{Document\}}

\emph{Summary: \{Summary\}}

\emph{Determine if the summary is factually consistent with the document provided above. A summary is factually consistent with the document if all information in the summary can be entailed (either stated or implied) by the document. Please start your answer with ``Yes.'' or ``No.'' Please briefly explain the reason within 50 words.}
\end{adjustwidth}

\paragraph{Prompting LLMs to generate explanations before providing a final answer results in no performance differences.} In a small-scale experiment, we observed that when we prompt the model to generate an explanation before providing the final answer, the response generated by the model tends to begin with a sentence such as \emph{``This sentence/summary is factually (in)consistent with the document''}, and the actual explanation begins after the second sentence. Since this is similar to starting the response with \emph{``Yes''} or \emph{``No''}, we chose the latter for simplicity.

\section{Extended Results: LLMs as Summarizers} \label{sec:extend-summarizer}

\subsection{Relevance and Completeness Evaluation} \label{sec:relevance-completeness}

As shown in the \emph{Rel.} column in Table~\ref{tab:summ_basic_stats}, each LLM's average relevance score is close to the maximum of 1 (see Section ~\ref{sec:human-eval} for more details). We conclude that \textbf{all LLMs are quite capable of focusing on the requested topics, with bigger models achieving slightly better performance}.\footnote{This is based on limited observation on open-source LLMs since the model size of GPT-3.5-Turbo is unknown.}

Next, we compare the models’ performance at covering the requested topic. It is worth noting that although smaller LLMs can achieve equivalent or even superior performance in summary completeness, the length of summaries generated by small LLMs is much higher. For example, Vicuna-7B achieves 0.72 in completeness on MeetingBank with an average summary length of 72 words. In contrast, GPT-3.5-Turbo achieves a comparable completeness score of 0.74 with more concise summaries (53 words). This trend also holds true for the three WizardLM models of varying sizes. The larger the WizardLM model size, the more capable the model is of either maintaining or covering more relevant information in the summary while making the summary shorter (WizardLM-13B vs. WizardLM-30B). Therefore, we conclude that \textbf{larger LLMs are more capable of generating information-dense summaries compared to smaller LLMs.}

\begin{table}
\small
\centering
\renewcommand{\tabcolsep}{0.85mm}
\begin{tabular}{lcccccccc}
\toprule
\multirow{2}{*}{\textbf{\begin{tabular}[c]{@{}l@{}}Summ.\\ Model\end{tabular}}} & \multicolumn{4}{c}{\textbf{MediaSum}} & \multicolumn{4}{c}{\textbf{MeetingBank}} \\
\cmidrule(r){2-5} \cmidrule(r){6-9}
 & \textbf{Len} & \textbf{\begin{tabular}[c]{@{}c@{}}Rel\\ {[}0,1{]}\end{tabular}} & \textbf{\begin{tabular}[c]{@{}c@{}}Cmp\\ {[}0,1{]}\end{tabular}} & \textbf{\begin{tabular}[c]{@{}c@{}}Err\\ \%\end{tabular}} & \textbf{Len} & \textbf{\begin{tabular}[c]{@{}c@{}}Rel\\ {[}0,1{]}\end{tabular}} & \textbf{\begin{tabular}[c]{@{}c@{}}Cmp\\ {[}0,1{]}\end{tabular}} & \textbf{\begin{tabular}[c]{@{}c@{}}Err\\ \%\end{tabular}} \\
\midrule
Vicuna-7B & 65 & 0.89 & 0.64 & 47.3 & 72 & 0.81 & 0.72 & 43.6 \\
Wizard-7B & 44 & 0.84 & 0.53 & 51.4 & 51 & 0.76 & 0.61 & 41.0 \\
Wizard-13B & 70 & 0.87 & 0.69 & 38.9 & 73 & 0.88 & 0.75 & 43.6 \\
Wizard-30B & 69 & 0.91 & 0.72 & 40.3 & 66 & 0.88 & 0.75 & 37.2 \\
GPT-3.5-Tb & 57 & 0.91 & 0.70 & 24.0 & 53 & 0.91 & 0.74 & 14.7 \\
\bottomrule
\end{tabular}
\caption{\textbf{Summarization model statistics on \textsc{TofuEval}.} For each model under evaluation, we include the human-evaluated completeness score (\emph{Cmp}), relevance score (\emph{Rel}) and percentage of summaries with at least one factual inconsistency (\emph{Err \%}) for each dataset. \emph{Wizard} is an abbreviation for WizardLM.} \label{tab:summ_basic_stats}
\end{table}

\subsection{Factual Consistency Evaluation} \label{sec:summ-factual}

\paragraph{Existing LLMs still make a considerable amount of factual errors.} In Table~\ref{tab:summ_basic_stats}, we show the percentage of summaries with at least one factually inconsistency for each model across datasets, according to our annotations. We find that out of all models we evaluated except GPT-3.5-Turbo, approximately 40--50\% of their summaries contain at least one factual inconsistency. Furthermore, \textbf{there is no direct positive correlation between a summary's length and the quantity of errors it contains}. For example, on MediaSum data, 38.9\% of WizardLM-13B's summaries were factually inconsistent with an average summary length of 70; whereas 24.0\% of GPT-3.5-Turbo's summaries are inconsistent while having a higher average length of 57. The computed Pearson correlation coefficient $\rho$ between models' length and proportion of inconsistent summaries is 0.18, with a $p$-value of 0.57, showing weak positive correlation.

\paragraph{Larger LLMs do not necessarily generate fewer factually inconsistent summaries.} As shown in Table~\ref{tab:summ_basic_stats}, when comparing the same model family, WizardLM-30B generates a slightly higher number of errors than WizardLM-13B on MediaSum, and WizardLM-13B generates more errors than WizardLM-7B on MeetingBank. Furthermore, while a larger model may have a lower error rate, the reduction may be minor. For example, WizardLM-30B's error rate is only 3.8\% lower than WizardLM-7B's on MeetingBank data. Comparing models from different families, the error rate of WizardLM-13B is the same as that of Vicuna-7B on MeetingBank.

\begin{table*}
\small
\centering
\begin{tabular}{clcccc|cccc}
\toprule
\multicolumn{1}{l}{\multirow{3}{*}{\textbf{\begin{tabular}[c]{@{}l@{}}Model\\ Type\end{tabular}}}} & \multirow{3}{*}{\textbf{\begin{tabular}[c]{@{}l@{}}Evaluation\\ Model\end{tabular}}} & \multicolumn{4}{c}{\textbf{Sentence-Level (FPR $\downarrow$)}} & \multicolumn{4}{c}{\textbf{Summary-Level (FPR $\downarrow$)}}  \\
\cmidrule(r){3-6} \cmidrule(r){7-10}
\multicolumn{1}{l}{} &  & \multicolumn{2}{c}{\textbf{MediaSum}} & \multicolumn{2}{c}{\textbf{MeetingBank}} & \multicolumn{2}{c}{\textbf{MediaSum}} & \multicolumn{2}{c}{\textbf{MeetingBank}} \\
\cmidrule(r){3-4} \cmidrule(r){5-6} \cmidrule(r){7-8} \cmidrule(r){9-10}
\multicolumn{1}{l}{} &  & \textbf{Main} & \textbf{Marginal} & \textbf{Main} & \textbf{Marginal} & \textbf{Main} & \textbf{Marginal} & \textbf{Main} & \textbf{Marginal} \\
\midrule
\multirow{4}{*}{\textbf{\begin{tabular}[c]{@{}c@{}}Non-\\ LLM\end{tabular}}} & SummaC-ZS  & 26.1 & 26.0 & 26.9  & \cellcolor{green3}20.2  & 51.9 & 51.8 & 35.6 & 38.1  \\
 & SummaC-CV & 47.2 & 40.2 & 27.9  & 30.6 & 53.5 & 46.9 & \cellcolor{green3} 24.3 & \cellcolor{green3}21.7 \\
 & QAFactEval & 33.2 & 22.7 &  30.1 & 29.2  & 35.4 & \cellcolor{green3}22.8 & 45.9 & 45.1 \\
 & AlignScore & 23.9 & 26.0 & 14.9  & 25.4  & 41.6 & 35.8 & 36.6 & 47.1 \\
\midrule
\multirow{2}{*}{\textbf{\begin{tabular}[c]{@{}c@{}}Prop.\\ LLM\end{tabular}}} 
 & GPT-3.5-Turbo & \cellcolor{green3}3.7 & \cellcolor{green3}14.6 & \cellcolor{green3} 13.4  & 48.4  & \cellcolor{green3}11.2 & 26.2 & 25.3 & 43.7  \\
 & GPT-4 & \cellcolor{green2}1.4 & \cellcolor{green2}5.7 & \cellcolor{green2}3.5  & \cellcolor{green2}6.7 & \cellcolor{green2}3.5 & \cellcolor{green2}12.5 & \cellcolor{green2}4.5 & \cellcolor{green2}7.8  \\
\bottomrule
\end{tabular}
\caption{\textbf{Sentence-level and summary-level false negative rate (FNR) for factual consistency evaluators on the test set of \textsc{TofuEval}} (a model incorrectly predicts that a summary or summary sentence contains an error).} \label{tab:FPR}
\end{table*}

\begin{table*}
\small
\centering
\begin{tabular}{clcccc|cccc}
\toprule
\multicolumn{1}{l}{\multirow{3}{*}{\textbf{\begin{tabular}[c]{@{}l@{}}Model\\ Type\end{tabular}}}} & \multirow{3}{*}{\textbf{\begin{tabular}[c]{@{}l@{}}Evaluation\\ Model\end{tabular}}} & \multicolumn{4}{c}{\textbf{Sentence-Level (FNR $\downarrow$)}} & \multicolumn{4}{c}{\textbf{Summary-Level (FNR $\downarrow$)}}  \\
\cmidrule(r){3-6} \cmidrule(r){7-10}
\multicolumn{1}{l}{} &  & \multicolumn{2}{c}{\textbf{MediaSum}} & \multicolumn{2}{c}{\textbf{MeetingBank}} & \multicolumn{2}{c}{\textbf{MediaSum}} & \multicolumn{2}{c}{\textbf{MeetingBank}} \\
\cmidrule(r){3-4} \cmidrule(r){5-6} \cmidrule(r){7-8} \cmidrule(r){9-10}
\multicolumn{1}{l}{} &  & \textbf{Main} & \textbf{Marginal} & \textbf{Main} & \textbf{Marginal} & \textbf{Main} & \textbf{Marginal} & \textbf{Main} & \textbf{Marginal} \\
\midrule
\multirow{4}{*}{\textbf{\begin{tabular}[c]{@{}c@{}}Non-\\ LLM\end{tabular}}} & SummaC-ZS  & 41.8 & 26.4 &  \cellcolor{green2}45.4 & 18.7  & \cellcolor{green2}22.8 & \cellcolor{green2}20.1 & 48.1 & \cellcolor{green2} 17.3  \\
 & SummaC-CV & \cellcolor{green2}17.8 & \cellcolor{green2}14.0 & 47.1 & \cellcolor{green3}14.9  & \cellcolor{green3}24.2 & \cellcolor{green2}20.1 & 71.0 & 32.6 \\
 & QAFactEval & 59.0 & 29.5 & 54.1  & 19.4  & 41.9 & 28.8 & \cellcolor{green3}43.9 & \cellcolor{green3}18.7 \\
 & AlignScore & \cellcolor{green3}37.9 & \cellcolor{green3}21.6 & 62.7  & 17.5  & 27.4 & \cellcolor{green2}20.1 & \cellcolor{green2}36.6 & \cellcolor{green2} 17.5 \\
\midrule
\multirow{2}{*}{\textbf{\begin{tabular}[c]{@{}c@{}}Prop.\\ LLM\end{tabular}}} 
 & GPT-3.5-Turbo & 73.0 & 47.7 & 48.4  & 21.6  & 69.6 & 42.2 & 48.4 & 24.9  \\
 & GPT-4 & 68.8 & 33.9 & \cellcolor{green3} 46.1  & \cellcolor{green2} 12.6  & 69.0 & 29.7 & 46.1 & 26.1  \\
\bottomrule
\end{tabular}
\caption{\textbf{Sentence-level and summary-level false positive rate (FPR) for factual consistency evaluators on the test set of \textsc{TofuEval}} (a model incorrectly predicts that a summary or summary sentence is correct).} \label{tab:FNR}
\end{table*}

\paragraph{Dataset affects model error rate.} As shown in Table~\ref{tab:topic_summ_stats}, models, on average, make more errors on MediaSum than on MeetingBank. The difference is more significant for the main topics and is magnified by the summary-level performance compared to the sentence level. This shows that there is a non-negligible impact of text distribution on the models' summarization performance. One hypothesis is that it is more challenging for models to generate factually consistent summaries related to a specific topic that requires aggregating and synthesizing information across conversational turns, and we find topic-related information tends to be more evenly distributed across conversational turns in MediaSum than in MeetingBank.

\section{Extended Results: LLMs as Evaluators} \label{sec:extend-llm-evaluator}

In addition to assessing evaluators' performance through balanced accuracy and recall, as detailed in Section~\ref{sec:llm-as-evaluators}, we also examine the reliability of evaluators in identifying factually inconsistent summary or summary sentences by analyzing the false positive rate (FPR) and false negative rate (FNR):
$$ \mathrm{FPR} = \frac{\mathrm{FP}}{\mathrm{FP} + \mathrm{TN}}, \quad \mathrm{FNR} = \frac{\mathrm{FN}}{\mathrm{FN} + \mathrm{TP}}.$$

On sentence-level evaluation, FPR indicates the rate at which an evaluator incorrectly predicts that a summary sentence contains an error when it is actually correct, where FP represents the number of false positives (factually consistent sentences incorrectly labeled as inconsistent) and TN represents the true negatives (factually consistent sentences correctly identified as such). A high FPR suggests that the evaluator is frequently flagging summary sentences as erroneous when they are factually consistent.

FNR represents the rate at which an evaluator incorrectly predicts that a summary sentence is correct when it actually contains an error, where FN stands for false negatives (factually inconsistent summary sentences incorrectly labeled as correct) and TP stands for true positives (factually inconsistency summary sentences correctly identified as such). A high FNR means the evaluator often overlooks errors in the summary sentences.

It is worth mentioning that FPR and FNR provide a more detailed breakdown of evaluators' performance captured by balanced accuracy (BAcc) in Table~\ref{sec:eval-performance}:
$$\mathrm{BAcc} = 1 - \frac{1}{2}(\mathrm{FPR} + \mathrm{FNR}).$$

\paragraph{Significant Test} The \colorbox{green2}{highlighted} performance is significantly better than the rest with p-value < 0.05 by a paired bootstrap test across all tables in this work.

\paragraph{LLM-based evaluators often overlook errors, while non-LLM-based factual consistency metrics tend to produce false alarms.} In line with the approach detailed in Section~\ref{sec:metric-error-analysis}, we show our findings for a subset of relatively strong evaluators in Tables~\ref{tab:FPR} and~\ref{tab:FNR}. The non-LLM-based metrics exhibit a significant issue with a high FPR. When these metrics signal a potential error, it necessitates a manual comparison between the summary sentence and the source document to verify its accuracy. This process results in a considerable amount of unnecessary effort. On the other hand, LLM-based evaluators display a higher FNR. While this might reduce the immediate workload, it introduces the risk of missing inconsistent sentences, which can be detrimental in the long run.

\paragraph{All evaluation models miss fewer errors for marginal-topic summary sentences, but LLM-based evaluators bring more false alarms when evaluating marginal-topic summaries.} As shown in Table~\ref{tab:FNR}, all models have a decreased FNR when evaluating summary or summary sentences from marginal topics.  Notably, this trend is more pronounced for LLM-based evaluators. For instance, LLMs show a substantial 20\% to 40\% decrease in sentence-level FNR, indicating their improved error detection capabilities. However, it appears that LLMs achieve this by identifying more summary sentences as factually inconsistent, leading to a higher FPR on marginal topics (Table~\ref{tab:FPR}). This is particularly noticeable in the case of GPT-3.5-Turbo.

\begin{table}
\small
\centering
\renewcommand{\tabcolsep}{1.2mm}
\begin{tabular}{clcccc}
\toprule
\multicolumn{1}{l}{\multirow{3}{*}{\textbf{\begin{tabular}[c]{@{}l@{}}Model\\ Type\end{tabular}}}} & \multirow{3}{*}{\textbf{\begin{tabular}[c]{@{}l@{}}Evaluation\\ Model\end{tabular}}} & \multicolumn{4}{c}{\textbf{Summary-Level (BAcc $\uparrow$)}} \\
\cmidrule(r){3-6}
\multicolumn{1}{l}{} &  & \multicolumn{2}{c}{\textbf{MediaSum}} & \multicolumn{2}{c}{\textbf{MeetingBank}} \\
\cmidrule(r){3-4} \cmidrule(r){5-6}
\multicolumn{1}{l}{} &  & \textbf{Main} & \textbf{Margin} & \textbf{Main} & \textbf{Margin} \\
\midrule
- & Baseline & 50.0 & 50.0 & 50.0 & 50.0 \\
\midrule
\multirow{4}{*}{\textbf{\begin{tabular}[c]{@{}c@{}}Non-\\ LLM\end{tabular}}} & SummaC-ZS & \cellcolor{green3}62.7 & 64.1 & \cellcolor{green3}58.1 & \cellcolor{green3} 72.4 \\
 & SummaC-CV & 61.2 & 66.5 & 52.4 & \cellcolor{green3}72.9  \\
 & QAFactEval & 61.4 & \cellcolor{green2}74.2 & 55.1 & 68.2  \\
 & AlignScore & \cellcolor{green2}65.5 & \cellcolor{green3}72.1 & \cellcolor{green2}63.4 & 71.8 \\
\midrule
\multirow{4}{*}{\textbf{\begin{tabular}[c]{@{}c@{}}Open\\ Source\\ LLM\end{tabular}}} & Vicuna-13B & 49.8 & 52.3 & 48.6 & 53.1  \\
 & Vicuna-33B & 50.0 & 50.5 & 50.8 & 47.6    \\
 & Wizard-13B & 52.0 & 52.3 & 47.3 & 51.9    \\
 & Wizard-30B & 50.0 & 51.8 & 50.0 & 51.2 \\
\midrule
\multirow{2}{*}{\textbf{\begin{tabular}[c]{@{}c@{}}Prop.\\ LLM\end{tabular}}} 
 & GPT-3.5-Turbo & 55.0 & 71.2 & 51.5 & 59.9   \\
 & GPT-4 & 58.2 & 68.5 & 56.6 & \cellcolor{green2}80.0  \\
\bottomrule
\end{tabular}
\caption{\textbf{Summary-level BAcc on the test set of \textsc{TofuEval} by \emph{directly evaluating on whole summaries}}. Directly predicting the factual consistency of summaries is worse than aggregating sentence-level factuality prediction results for all models (Table~\ref{sec:eval-performance}).} \label{sec:summ-level-eval-performance}
\end{table}

\section{\textsc{TofuEval} Annotation Instructions} \label{sec:annotation-instruction}

We separated the human evaluation work for the \textsc{TofuEval} benchmark into two tasks due to the workload. Each task was annotated by a different group of annotators. There are 300 annotations for each of the two tasks (2 datasets $\times$ 50 documents $\times$ 3 topics). The first task (Task 1) consisted of topic categorization, factual consistency evaluation, and relevance evaluation. The second annotation task (Task 2) involved completeness evaluation.

\subsection{Task 1} \label{sec:Annotation-task-1}

\paragraph{Topic Categorization} is defined as $T_{\mathrm{topic}}$ (dialogue document, topic) $\to$ \{main, marginal, irrelevant\}. We defined the categories as follows:

\emph{Main Topic} refers to the central information in a document that is under discussion or is presented in the document. The main topics are often what the document is primarily about, and understanding them is critical to understanding the overall idea of the document.

\emph{Marginal Topic} refers to information in a document that is not the main focus of the document but is still part of the context. These topics are typically less prominent or less extensively explored than the main topics. They may contribute to the overall context, provide additional information, or enhance understanding of the main topics, but they are not the primary focus.

\emph{Irrelevant Topic} refers to information in a document that is not directly related to the subject or purpose of the document. Such topics might not contribute to the main topic(s) or objective of the document and can be seen as a diversion or distraction from the main topics at hand.

See Section~\ref{sec:agreement} for information about our post-processing of topic categories, in which we merged marginal and irrelevant topics together after data collection.

\paragraph{Factual Consistency Evaluation} is defined as $T_{\mathrm{fact}}$(dialogue document, sentence) $\to$ \{consistent, inconsistent\}, where a factually
consistent sentence is a sentence that is entailed (either stated or implied) by the document.

\begin{table}
\small
\centering
\renewcommand{\tabcolsep}{1.25mm}
\begin{tabular}{clcccc}
\toprule
\multicolumn{1}{l}{\multirow{3}{*}{\textbf{\begin{tabular}[c]{@{}l@{}}Model\\ Type\end{tabular}}}} & \multirow{3}{*}{\textbf{\begin{tabular}[c]{@{}l@{}}Evaluation\\ Model\end{tabular}}} & \multicolumn{4}{c}{\textbf{Summary-Level (BAcc $\uparrow$)}} \\
\cmidrule(r){3-6}
\multicolumn{1}{l}{} &  & \multicolumn{2}{c}{\textbf{MediaSum}} & \multicolumn{2}{c}{\textbf{MeetingBank}} \\
\cmidrule(r){3-4} \cmidrule(r){5-6}
\multicolumn{1}{l}{} &  & \textbf{Main} & \textbf{Margin} & \textbf{Main} & \textbf{Margin} \\
\midrule
- & Baseline & 50.0 & 50.0 & 50.0 & 50.0 \\
\midrule
\multirow{4}{*}{\textbf{\begin{tabular}[c]{@{}c@{}}Open\\ Source\\ LLM\end{tabular}}} & Vicuna-13B & 51.7 & 49.2 & 49.9 & 49.6  \\
 & Vicuna-33B & 54.4 & 56.0 & 54.6 & 54.0    \\
 & Wizard-13B & 56.8 & 57.1 & 56.2 & 60.1    \\
 & Wizard-30B & 56.0 & 57.7 & 54.9 & 56.1 \\
\midrule
\multirow{2}{*}{\textbf{\begin{tabular}[c]{@{}c@{}}Prop.\\ LLM\end{tabular}}} 
 & GPT-3.5-Turbo & \cellcolor{green3} 60.1 & \cellcolor{green3} 64.3 & \cellcolor{green3} 61.9 & \cellcolor{green3} 61.5   \\
 & GPT-4 & \cellcolor{green2}64.2 & \cellcolor{green2}78.7 & \cellcolor{green2}75.9 & \cellcolor{green2}83.9  \\
\bottomrule
\end{tabular}
\caption{\textbf{Summary-level BAcc on the test set of \textsc{TofuEval} for the \emph{\textsc{Exp} setting}}, where we ask a model to provide explanations for its decisions in addition to providing binary judgments (Section~\ref{sec:predict-label}). Summary-level labels are obtained by aggregating sentence-level labels. Non-LLMs do not provide explanations.}\label{sec:summ-exp-level-eval-performance}
\end{table}

Note that we conducted this task at the sentence level, with annotators having access to the entire summary. If any sentence is labeled as factually inconsistent, annotators are asked to explain their reasoning in natural text.

\paragraph{Relevancy Evaluation} is defined as $T_{\mathrm{rel}}$ (dialogue document, topic, summary) $\to$ \{1, 2, 3, 4, 5\}. We defined the 1-5 Likert scale as follows.

5-Excellent: The summary does not contain non-topic related content; 4-Very Good: The summary contains a small amount of non-topic related content; 3-Good: Half of the summary is off-topic. The content is somewhat balanced between topic-related and non-topic related content; 2-Fair: More than half of the summary is off-topic, but there is still topic related content; 1-Poor: The summary is composed of non-topic related content.

\begin{figure*}
    \centering
\includegraphics[width=\linewidth, trim=0mm 00mm 00mm 0mm,clip]{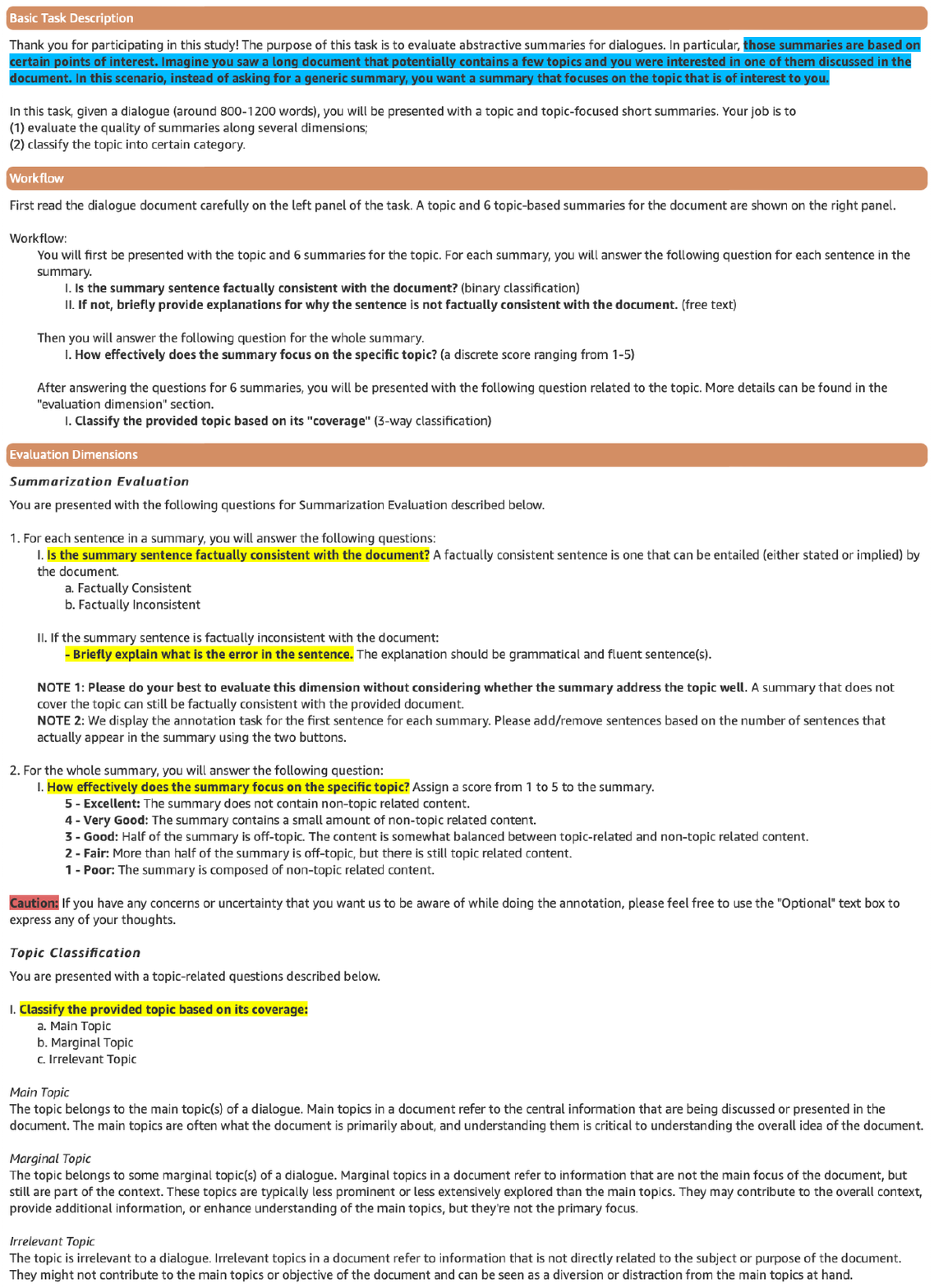}
    \caption{Screenshot of annotation interface for (1) factual consistency evaluation; (2) relevance evaluation; and (3) topic categorization.}
    \label{fig:consistency-1}
\end{figure*}

\begin{figure*}
    \centering
\includegraphics[width=\linewidth, trim=0mm 00mm 00mm 0mm,clip]{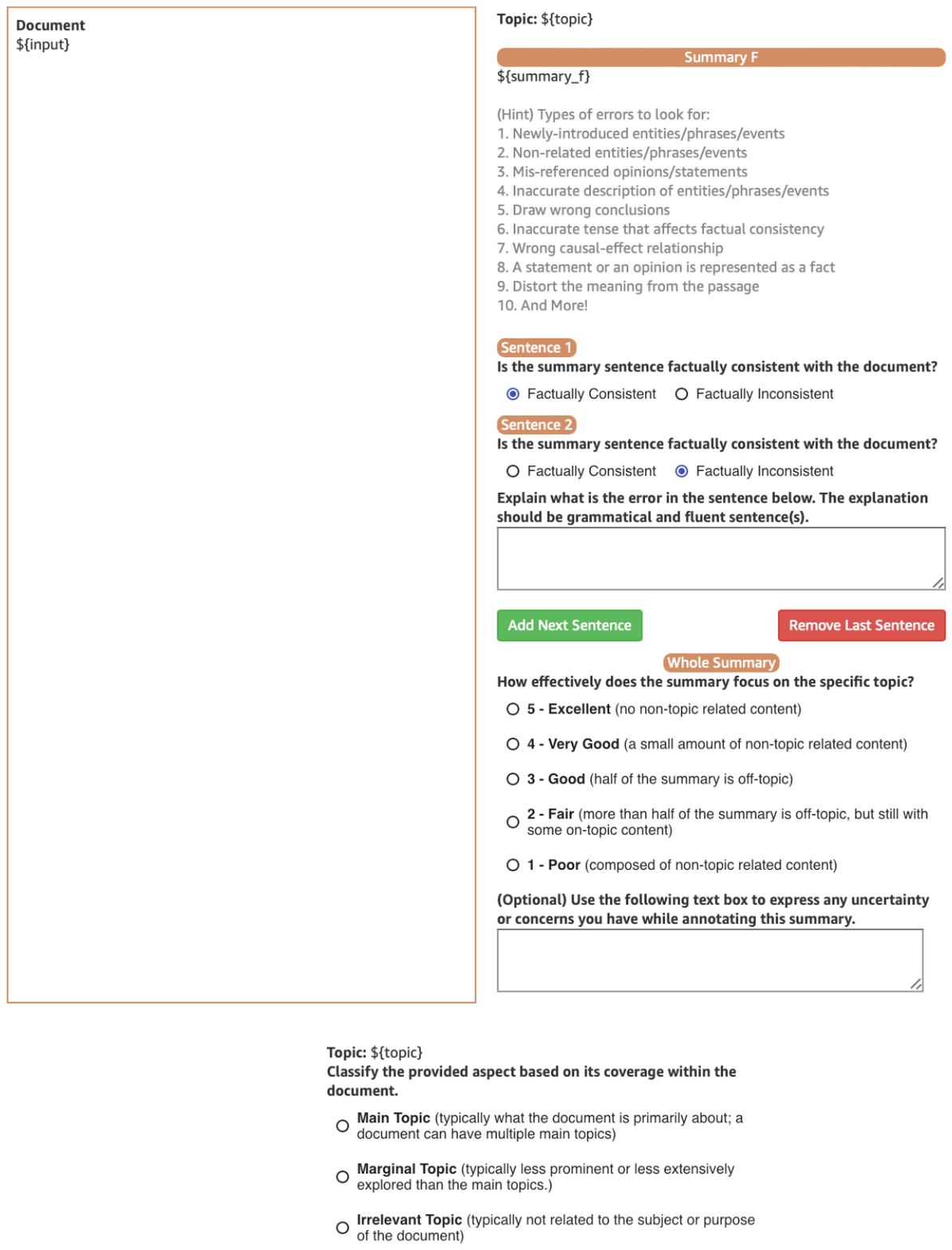}
    \caption{(Continue) Screenshot of annotation interface for (1) factual consistency evaluation; (2) relevance evaluation; and (3) topic categorization. We provide some possible error categories that annotators could look for during the annotation. \textbf{Note that we devised these error categories during the practice session and this is not the final version of out error taxonomy mentioned in Section~\ref{sec:llm-summarizer}}.}
    \label{fig:consistency-2}
\end{figure*}

See Section~\ref{sec:agreement} for an explanation of how we merged scores ($\{1, 2\} \rightarrow 0$, and $\{3, 4, 5\} \rightarrow 1$) to improve annotation agreement. The annotation guidelines and interface for Task 1 can be found in Figure~\ref{fig:consistency-1} and ~\ref{fig:consistency-2}.

\paragraph{Two Pass Annotations} To ensure the quality of our collected annotations, a subset of the annotation task was completed by two separate annotators.  We used the feedback from these two-pass annotations to identify any ambiguities or issues in the annotation guidelines and make necessary revisions. For more details on how we enhanced consensus among annotations, please refer to Section~\ref{sec:quality-control}.

\subsection{Task 2} \label{sec:Annotation-task-2}

\paragraph{Completeness Evaluation} Inspired by the Pyramid method \cite{nenkova-passonneau-2004-evaluating}, we first asked an annotator to write down $n$ key points in grammatical sentences: $T_{\mathrm{keys}}$(dialogue document, topic) $\to$ $K$, where $K=\{k_1$, ..., $k_n$\}. These key points were supposed to be what the annotator thought an ideal summary covering a given topic should include. For each key point $k_i$, we then asked the annotator whether a given summary contains $k_i$, \emph{i.e.}, $\mathds{1}$(summary, $k_i$). After the completeness annotation was finished, we calculated the completeness score in the following way: $\frac{1}{n} \sum_{i=1}^{n} \mathds{1}$(summary, $k_i$).

\begin{figure*}
    \centering
\includegraphics[width=\linewidth, trim=0mm 00mm 00mm 0mm,clip]{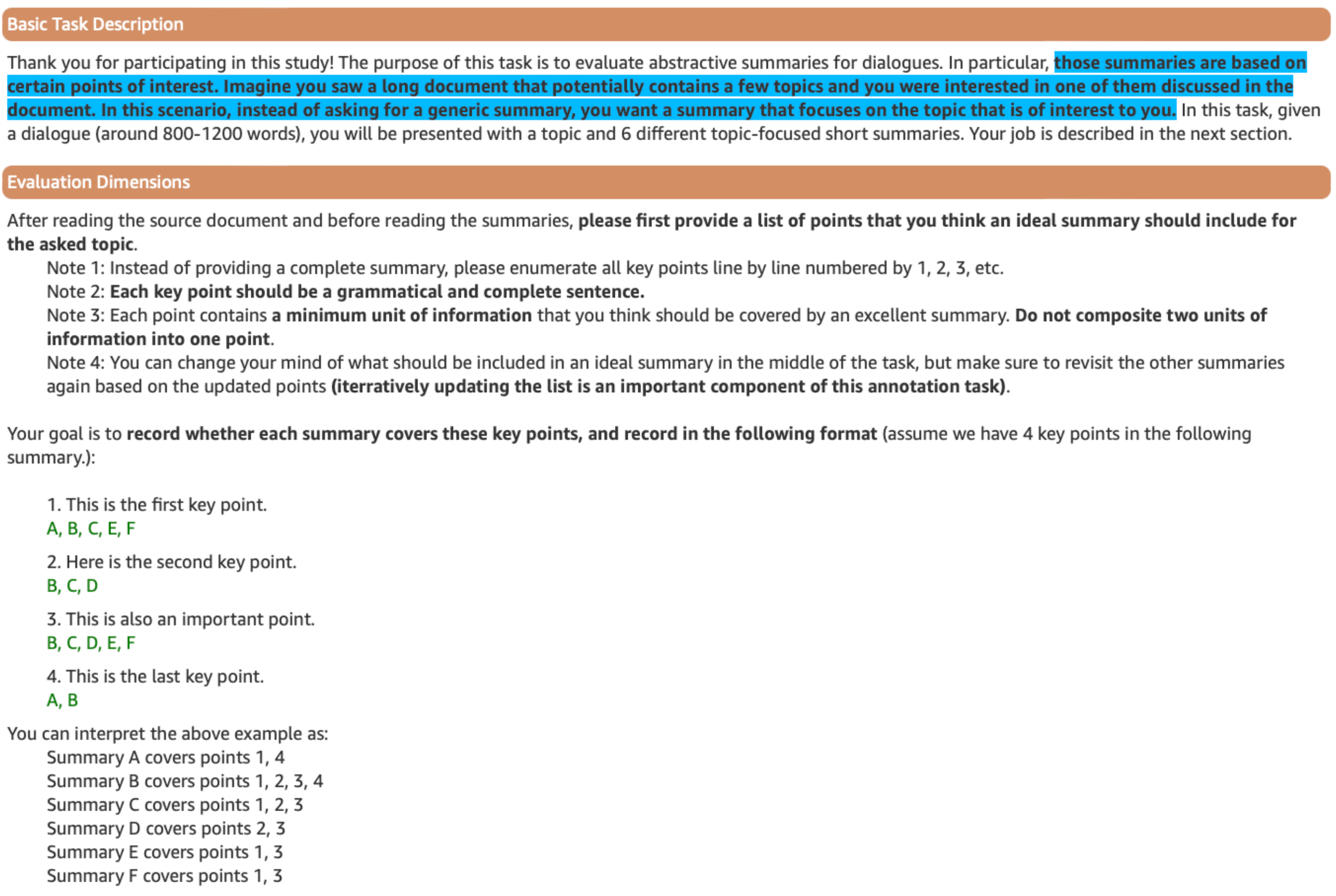}
    \caption{Screenshot of annotation interface for completeness evaluation. This is a separate annotation task from the previous one due to the workload.}
    \label{fig:completeness-1}
\end{figure*}

\begin{figure*}
    \centering
\includegraphics[width=\linewidth, trim=0mm 00mm 00mm 0mm,clip]{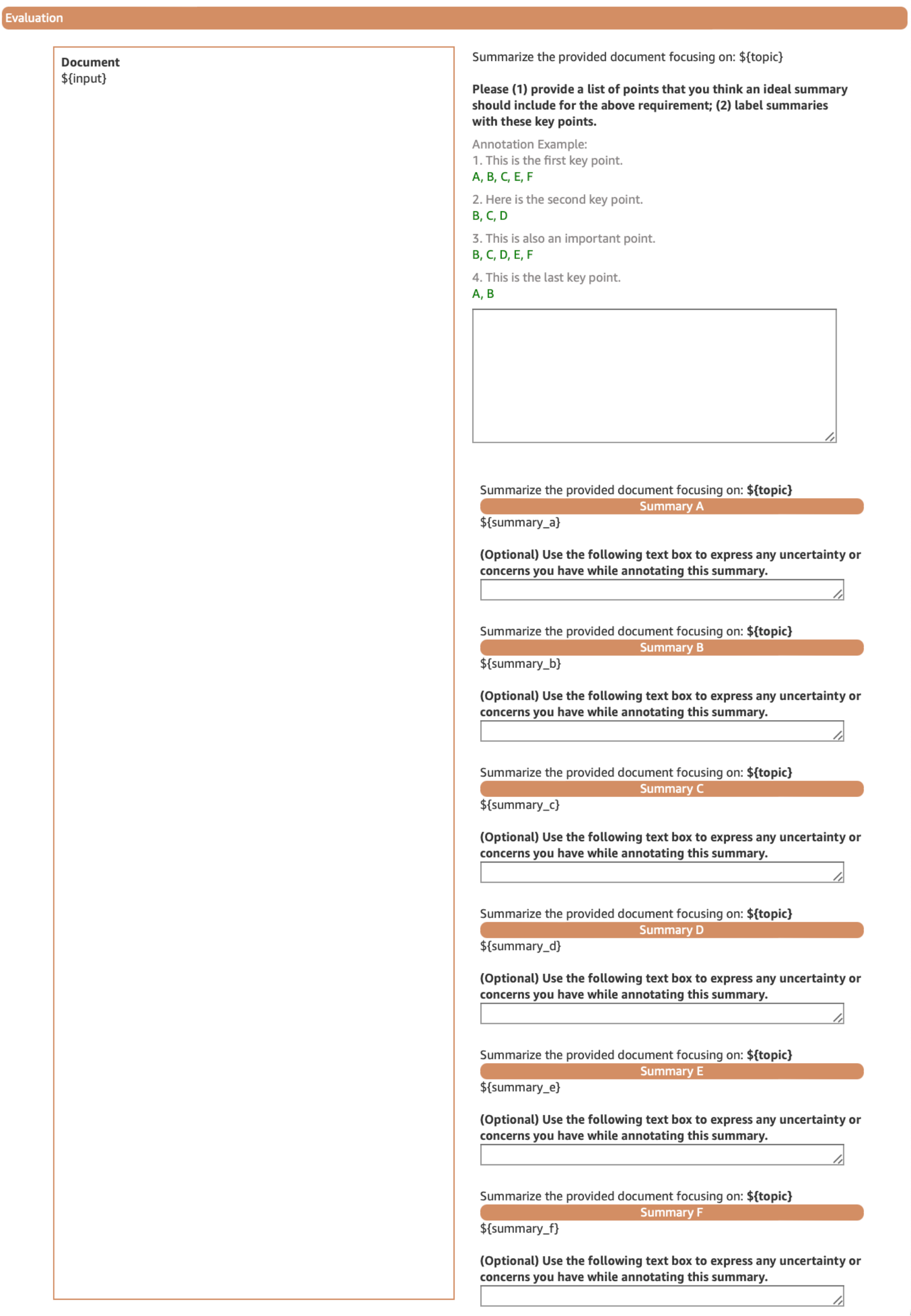}
    \caption{(Continue) Screenshot of annotation interface for completeness evaluation.}
    \label{fig:completeness-2}
\end{figure*}

Since it is impractical to consolidate key points from multiple annotators, we only utilized one annotator for each task and provided the written key points for reproducibility. The annotation guidelines and interface for Task 2 can be found in Figure~\ref{fig:completeness-1} and ~\ref{fig:completeness-2}.

We would like to note that we found that \textbf{annotating completeness using a 1-5 Likert scale diminished the quality of the results}. Without asking annotators to explicitly write down the important key points of a dialogue, we found that the order in which summaries were presented to annotators had a noticeable impact on what they considered important to include in an ideal summary. This effect occurred because the annotator's perception of what should be in a summary may change based on recently-read summaries. As a result, annotators needed to revise their annotations for summaries they had already assessed, toggling back and forth to align evaluations they had already finalized evaluations with their evolving opinions. This situation may result in an increase in errors and a reluctance or disinclination to amend prior responses, ultimately leading to unreliable annotations. Therefore, we encourage future work to evaluate the completeness of summaries by adopting our approach.

\begin{figure*}
    \centering
\includegraphics[width=\linewidth, trim=0mm 0mm 00mm 0mm,clip]{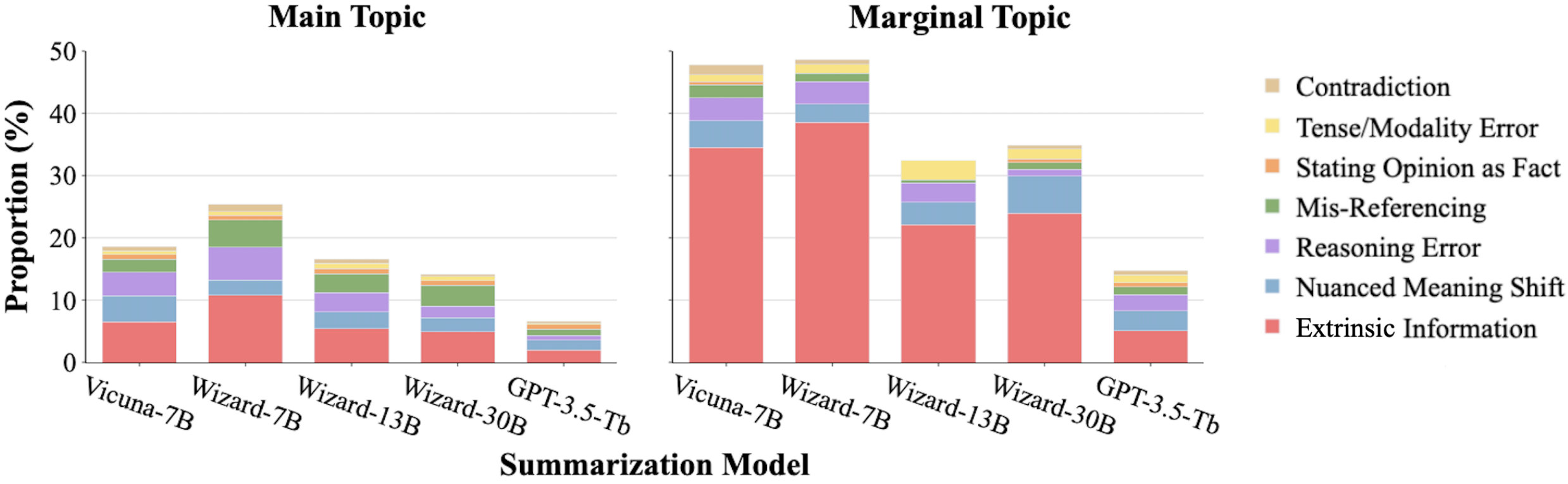}
    \caption{Error distributions over all summary sentences for each summarizer for main/marginal topics.}
    \label{fig:error_dist_plot_appendix}
\end{figure*}

\subsection{Quality Control} \label{sec:quality-control}

\paragraph{Arranging the Practice Session} We developed the annotation guidelines and arranged a practice session for the annotation tasks with the assistance of a select group of professional linguistic data annotators. During our preparation of the practice session, we refined the guidelines and answers to the aforementioned dimensions, emphasizing precautions we wanted the annotators to consider during annotation. We engaged in discussions with the group to reach a consensus on these answers. Using the finalized guidelines, all professional linguistic data annotators participated in a practice session to familiarize themselves with the tasks and calibrate their evaluations.

\paragraph{Providing Customized Feedback} After the practice session, we held multiple rounds of pilot annotations. In each round, every annotation task was undertaken by two annotators so that we could calculate inter-annotator agreement to ensure that we could maintain or improve the agreement rate. After each round, we compared the annotations, provided individualized feedback to each annotator, and refined the annotation guidelines. Once we achieved a converging annotation agreement rate, we proceeded with the remaining annotation tasks.

\paragraph{Annotation Efforts} The average time spent per annotation on the two annotation tasks was 36 and 24 minutes, respectively. All of the annotators involved in the human evaluation tasks presented in the current work have native-level proficiency in English, and they are compensated a competitive hourly rate that is benchmarked against similar roles in their country of residence.

\subsection{Inter-Annotator Agreement} \label{sec:agreement}

We use Cohen's Kappa $\kappa$ \cite{McHugh2012} to measure human annotation agreement on tasks where we received two annotation passes (Task 1). Below, we provide $\kappa$ for each evaluation dimension in Task 1 as well as the post-processing steps we undertook to improve annotation agreement.

\paragraph{Topic Categorization and Mapping} Agreement was moderate for both datasets (for MediaSum, $\kappa$ = 0.47; for MeetingBank, $\kappa$ = 0.53). Sinceonly 2\% of topics in the benchmark were annotated as irrelevant and we observed that annotators frequently label ``marginal'' and ``irrelevant'' interchangeably in these topics, we merged \{marginal, irrelevant\} $\to$ \{marginal\} after the annotations were completed.

\paragraph{Factual Consistency} We achieved $\kappa$ = 0.42 and 0.34 for evaluations on MediaSum and MeetingBank summaries, respectively, showing fair to moderate agreement. Based on human-written explanations, we found that the primary disagreements occurred when the annotators erroneously labeled sentences as factually inconsistent, when the summary was incomplete, when summary sentences did not directly address the topic, or when other situations arose that should not be considered under this evaluation dimension.

Given these findings, after we collected all annotations, the authors of the current work performed a \emph{second-stage annotation review} to eliminate any false positive labels from the benchmark dataset based on written explanations. Summarization models were hidden during this review process to bar against any unconscious bias. We want to emphasize the reliability of our second-stage review process. It was conducted using our definitions for factual consistency along with careful consideration of the explanations provided by the annotators. The reviewers discussed challenging cases to ensure alignment, and all decisions were reviewed by another reviewer. We believe this process substantially improved the quality of the dataset.

\paragraph{Relevance} Unlike the factual consistency evaluation in which explanations were elicited, we found it challenging to achieve high agreement on relevance evaluation even after a few rounds of targeted feedback. We ultimately decided to bin the scores as follows: $\{1, 2\} \rightarrow 0$, and $\{3, 4, 5\} \rightarrow 1$. With this grouping, $\kappa$ = 0.25 for MediaSum summaries and 0.37 for MeetingBank summaries, which we consider to be fair agreement. We evaluated relevance on main-topic summaries only since topic-related content may not exist for marginal topics.

\paragraph{Writing explanations is helpful for providing feedback and improving annotation consensus.} We observed that the value of $\kappa$ for these three dimensions increased from 0.1 to 0.3 after the annotators received additional training. Notably, we found that improving annotation agreement is relatively straightforward for the dimension of factual consistency; based on the explanations written by the annotators, we can easily grasp the annotator's reasoning, allowing us to identify shortcomings, whether in their reasoning or in our annotation guidelines. This further enabled us to provide targeted feedback and incorporate clarifications to the guidelines, ultimately improving annotation agreement. For topic categorization and relevance evaluation, once a pilot round was completed, we asked annotators to share their thoughts and uncertainties on examples where we observed significant disagreement. We then refined the annotation guidelines based on their feedback.

\section{Error Type Curation and Annotation} \label{sec:err-type-anno}

\subsection{Error Type Curation}

Due to the complexity of the dialogues in \textsc{TofuEval}, we curated the following error types with professional linguistic data annotators (all coauthors of the current work) based on the explanations written by the annotators for factually inconsistent sentences. Note that our error taxonomy is strongly influenced by that from \citet{tang-etal-2022-confit}, which was initially proposed for short dialogues. A concise description of our error taxonomy with illustrated examples is provided in Figure~\ref{fig:error_type}.

\paragraph{Extrinsic Information Error} Similar to \citet{maynez-etal-2020-faithfulness}, we define the error as follows: the summary sentence contains new information that is not from the source document and cannot be verified from the source document. If the new information is not from the source document but is everyday commonsense knowledge, we do not label the sentence as containing an error.

\paragraph{Misreferencing Error} The summary sentence refers to an entity (e.g. as a noun phrase in subject/object position) that is grounded in the source document, but the sentence attributes a property or event to this entity; this wrong property or event is grounded in the source document but is attributed to a different entity in the source document.

\paragraph{Stating Opinion-as-Fact Error} The summary sentence presents a proposition as fact (i.e. there are no uncertainty modals or adverbs like \textit{might} or \textit{probably} in the sentence) when the proposition is presented as someone's opinion in the source document. 

\paragraph{Reasoning Error} The summary sentence makes wrong inferences (e.g. it contains error in arithmetic calculation, it draws incorrect relations between A and B, or it makes wrong extrapolations) based on premises or pieces of evidence that \emph{are} grounded in the source document. (Note that if the evidence for the inference is not grounded in the source document, that would be considered an Extrinsic Information Error).

\paragraph{Tense/Aspect/Modality Error} The summary sentence uses the wrong tense (e.g. past tense in the source document but future tense in the summary sentence), aspect (e.g. progressive in the source document but perfective in the summary sentence), or modality (e.g. epistemic possibility modal \textit{might} in the source document but epistemic necessity modal \textit{must} in the summary sentence).

\paragraph{Contradiction Error} The summary sentence fully contradicts the source document, either due to erroneous presence or absence of negation or due to the use of an antonym of a word used in the source document.

\paragraph{Nuanced Meaning Shift Error} The summary sentence alters the meaning of certain sentences in the source document by using paraphrasing with words or phrases that are associated with different senses (\emph{e.g.} paraphrasing "make a recommendation" to "make a request").

\paragraph{Others} In our annotation process, we intentionally introduced an \emph{other} error category. However, as no sentence was found to be factually inconsistent within this category, we have chosen not to include it in our error taxonomy.

\subsection{Error Type Annotation}

With the curated error taxonomy, the professional linguistic data annotators in the author list assigned one or more error types to all factually inconsistent sentences in \textsc{TofuEval} by referring to the summaries and the explanations provided by the annotators. The source documents were referred to if we could not identify the error type based on human explanations.

We conducted four rounds of pilot studies for the error type assignment task. The first two rounds were used to finalize the error taxonomy we modified from \citet{tang-etal-2022-confit}, while the following two rounds were dedicated to calibrating our error type categorization and improving our agreement rates (in cases of unresolved disagreement, we took the majority vote to arrive at the final error category or categories). For each pilot round, annotators assigned error types to the same set of sentences. We achieved a Fleiss' Kappa score of \cite{Fleiss1971} 0.78 after the final pilot round, indicating substantial agreement, and we proceeded with the remaining error type categorizations.

\section{Performance in Generating explanations} \label{sec:exp-eval}

It has been observed that LLMs can generate critiques of model outputs, in some cases leading to enhanced output quality \cite{madaan2023selfrefine, saunders2022self}. We now investigate whether LLMs are capable of generating correct explanations for factually inconsistent sentences.\footnote{Since we observed that all tested LLMs perform worse at binary factual error detection when prompted to consider a whole summary rather than an isolated summary sentence, we only evaluated the quality of the LLM-generated explanations at the sentence level.} In particular, we focused on examples where both humans and LLMs labeled the summary sentence as factually inconsistent, and we examine whether LLMs can generate correct explanations of the factual inconsistencies in these cases.

\begin{table}
\small
\centering
\renewcommand{\tabcolsep}{1.9mm}
\begin{tabular}{lclc}
\toprule
\multicolumn{2}{c}{\textbf{Open-Source LLM}} & \multicolumn{2}{c}{\textbf{Prop. LLM}} \\
\cmidrule(r){1-2} \cmidrule(r){3-4}
\textbf{Model} & \multicolumn{1}{l}{\textbf{Acc (\%)}} & \textbf{Model} & \textbf{Acc (\%)} \\
\midrule
Vicuna-13B & 45 &  &  \\
Vicuna-33B & 40 &  &  \\
WizardLM-13B & 60 & GPT-3.5-Turbo & 50 \\
WizardLM-30B & 55 & GPT-4 & \textbf{80} \\
\bottomrule
\end{tabular}
\caption{\textbf{Percentage of correct explanations across models on \textsc{TofuEval}.} We show \textbf{Human}'s evaluation results on a small sampled set of 20$\times$6=120 explanations where both human annotations and models predict that the summaries contain errors from the main-topic set, where we have more diverse error types.} \label{sec:exp-eval-by-human}
\end{table}

\paragraph{Human Evaluation} We randomly sampled 20 summary sentences to conduct a small-scale human evaluation task. Sentences were only selected if one of the eight LLM-based evaluators and a human annotator labeled the sentence as factually inconsistent. We only sampled from main-topic summary sentences since they contain a more diverse range of error types (Figure~\ref{fig:error_dist_plot}). Three authors of this work manually evaluated the quality of the model-generated explanations by comparing them to human-written explanations. Specifically, humans were provided with (1) a whole summary; (2) a factually inconsistent sentence in the summary; (3) a human-written explanation for the sentence; and (4) a model's explanation. We asked annotators to perform a binary classification task in which they determined whether the model-provided explanation was supported by or equivalent to the human-written explanation. We optionally provided the source document to the annotators in case they needed more context. The models that generated these explanations were hidden from the annotators.

It is possible that a factually inconsistent summary sentence could have semantically different but valid explanations. This could potentially impact the quality of our manual analysis if a model provides a reasonable explanation that is considered incorrect simply because the explanation does not resemble the one provided by an annotator. To quantify the potential impact of this, for tasks where we had completed two rounds of annotation (see Section~\ref{sec:human-eval} for more details), we compared the explanations generated by two annotators when they both identified a sentence as factually inconsistent. This investigation revealed that the explanations written by both annotators were semantically equivalent over 95\% of the time, suggesting that \textbf{the alternative model-generated explanations should not require much concern in \textsc{TofuEval}}.

We obtained a Fleiss Kappa score \cite{Fleiss1971} of 0.65, indicating substantial agreement. We took the majority vote to obtain the final label for each model-generated explanation (supported or not supported by the human explanation) and calculated the explanation accuracy for each LLM-based evaluator using the finalized labels. The result is provided in Table~\ref{sec:exp-eval-by-human}, where it can be observed that GPT-4 is capable of providing correct explanations 80\% of the time when it identifies that a summary sentence is factually inconsistent with the source document. Other models provide accurate explanations about half of the time without significant differences between the models.

\section{Computing Infrastructure}

For inference on the proprietary models, we used the official APIs. For LLMs with 7B and 13B parameters, we utilized a cluster of four Tesla V100-SXM2 GPUs, each with 16GB memory. For the larger 30B and 33B LLMs, we used four NVIDIA A100-SXM4 GPUs, each with 40GB of memory. For non-LLM-based models, we used a single Tesla V100-SXM2 GPU.

For API calls, we use \texttt{gpt-3.5-turbo-0613} for GPT-3.5-Turbo; \texttt{gpt-4-0613} for GPT-4.

\begin{table*}
\begin{tabular}{p{\linewidth}}
MediaSum - Document ID: CNN-25553 \\
\toprule
\textbf{CAROL LIN, CNN ANCHOR:} Well, the government is also about to tell us what airline passengers already know: that the service is less than perfect, even after the nation's carriers promised to upgrade service. A report comes from the Transportation Department's inspector general in less than four hours. And we are joined by a travel expert, Terry Trippler, in Minneapolis. Terry, I wonder if you've been flying lately because you got a chance at least to peak at the preliminary reports. What are we likely to hear? \\
\textbf{TERRY TRIPPLER, TRAVEL EXPERT:} I think what we're going to see is something similar to the interim report that came out in June of last year: improvement, but a long way to go. And I think that's what we're going to have happen again. We're going to see this noon. \\
\textbf{LIN:} Well, let's touch on at least some of the promises that the airlines said that they would try to work on. For example, when I go ahead and I book my ticket, am I guaranteed that the airline is going to quote me the cheapest fair? \\
\textbf{TRIPPLER:} That's a tough one, Carol. They have promised to do that. Some of the airlines are making pretty good on that promise. Other ones aren't doing too well. Basically, where the problem lies is in these last-minute Internet fares that they have that -- there are some passengers claim they're not being told about. So there needs to be some improvement on that area. \\ 
\textbf{LIN:} All right. Well, what are -- are they going to be able to tell me -- or will they tell me if the flight is oversold as I book my seat? \\ 
\textbf{TRIPPLER:} If you ask, from what I gather, they are telling you if the flight is oversold. What we find happening on this one is, once the passenger is finding the flight is oversold, they book on that flight because they want to get voluntarily bumped and get the miles and the money and the meals, etcetera. So that one sort of backfired on them. \\
\textbf{LIN:} All right, let's say they lose my luggage. How long is it going to take for me to get it back these days? \\
\textbf{TRIPPLER:} They claim they'll do it in 24 hours. Luggage complaints are up. And, of course, we recently all have seen the film of where the luggage handlers were playing basketball with people's packages, his luggage. That did not help. Complaints are up. They're going to have to do a better job on luggage. \\
\textbf{LIN:} All right, well, also, I find myself more often than not sitting on a plane, getting ready to taxi the runway, and suddenly everything comes to a halt. And I'm told that it is problems with air traffic control or something that really doesn't mean much to me as a passenger. Am I going to be told, or should I be told -- am I being told why I'm being delayed specifically? \\
\textbf{TRIPPLER:} Well, they say they are telling you. And here's where the big problem lies. And these are the complaints that I am receiving by the literally hundreds per day in e-mails. People feel they're not being told the truth. They're not being told before they board the aircraft that there's a possibility that they'll be delayed. I mean, people are boarding aircraft -- I did, I boarded one, went out and sat at the end of the tarmac. I was there long enough to qualify to vote in that precinct. They've got to do a better job on this. Get out of the gate. Get off the ground. \\
\textbf{LIN:} So if they're not quite meeting the promises that they said that they would keep to Congress, does this mean that Congress will or should go ahead with legislation to force the airlines to give better service? \\
\textbf{TRIPPLER:} Well, Carol, I think that, as soon as this report is issued, that we're going to have people lined up all the way around the Capitol to file a bill for passenger-rights legislation. Already we have one or two of the 1999 bills that have been resurrected. Those were bills that were put aside when the airlines promised to do a better job. Yes, I think the airlines, unfortunately -- and I'm a free-enterprise person -- but, unfortunately, I think the airlines have forfeited their right to operate without some type of government intervention. I think it's inevitable. It's got to happen. \\
\bottomrule
\end{tabular}
\caption{A dialogue example with generated summaries and human annotations in \textsc{TofuEval} (part 1). We show all three generated topics for the dialogue in the table. For each topic, we show one of the five generated summaries with human-written explanations and {\color{blue1}{error types}} for {\color{red3}{factually inconsistent sentences}}.} \label{tab:example-1}
\end{table*}

\begin{table*}
\begin{tabular}{p{\linewidth}}
MediaSum - Document ID: CNN-25553 \\
\toprule
\textbf{LIN:} Well, and some airlines are going to say that so many factors are out of their control: like weather and now labor problems. Delta's pilots are expected to announce today what they voted on Friday: whether they're actually going to go on strike. \\
\textbf{TRIPPLER:} That's right. And we're talking Northwest now, Delta. Before the year's over, we're going to be talking to American and United. And I disagree with the airlines there. I believe those are in their control. Weather, I understand. Labor: Come on, airlines, let's get it together. \\
\textbf{LIN:} All right, thank you very much for that preview. It really makes you want to hit the road, doesn't it?
\textbf{LIN:} Thanks, Mr. Trippler. \\
\textbf{TRIPPLER:} Thank you. \\
\\

\multicolumn{1}{c}{\begin{tabular}[c]{@{}c@{}}\textbf{Topic:} Airlines' commitment to service improvements (main)\end{tabular}} \\
\textbf{Summary:} \textbf{[1]} The airlines promised service improvements but have not fully delivered. {\color{red3}\textbf{[2]}} Complaints remain on inaccurate fare quotes, overbooking, delayed luggage return and lack of communication during delays. {\color{red3}\textbf{[3]}} Government intervention may be needed as airlines have forfeited the right to self-regulate. \\
\\
\textbf{Explanation for} {\color{red3}\textbf{[2]}}: Re: fare quotes, the complaint is about price transparency rather than price accuracy. {\color{blue1}{Nuanced Meaning Shift}} \\
\textbf{Explanation for} {\color{red3}\textbf{[3]}}: It is the opinion of Trippler, and not a fact, that the airlines have ``forfeited'' the right to self-regulate. {\color{blue1}{Stating opinion as fact}}\\
\\
\multicolumn{1}{c}{\begin{tabular}[c]{@{}c@{}}\textbf{Topic:} Factors affecting airlines' control over service quality (main)\end{tabular}} \\
\textbf{Summary:} \textbf{[1]} The government is set to release a report on the service quality of airlines, with improvements needed in areas such as quoting the cheapest fare, informing passengers about oversold flights, handling luggage complaints, and providing accurate information about delays. \textbf{[2]} The report may lead to legislation for passenger-rights and government intervention due to the airlines' failure to meet promises. {\color{red3}\textbf{[3]}} Some airlines argue that factors like weather and labor problems are beyond their control, but experts disagree. \\
\\
\textbf{Explanation for} {\color{red3}\textbf{[3]}}: Experts don't disagree that e.g. weather is beyond the airlines' control. {\color{blue1}{Contradiction}}\\
\\
\multicolumn{1}{c}{\begin{tabular}[c]{@{}c@{}}\textbf{Topic:} Passengers' rights and legislation (main)\end{tabular}} \\
\textbf{Summary:} {\color{red3}\textbf{[1]}} The government report will show airlines haven't fully delivered on service promises. {\color{red3}\textbf{[2]}} Passengers complain of inaccurate fare quotes, lack of info on delays/oversold flights, mishandled baggage. {\color{red3}\textbf{[3]}} Congress may pass legislation to mandate passenger rights as airlines have forfeited self-regulation. \\
\\
\textbf{Explanation for} {\color{red3}\textbf{[1]}}: It is not stated in the document whether the report will actually show that airlines haven't fully delivered on service promises. It is the travel expert's opinion. {\color{blue1}{Stating opinion as fact}} \\
\textbf{Explanation for} {\color{red3}\textbf{[2]}}: It is not about inaccurate fare quote but not getting the cheapest quote. {\color{blue1}{Nuanced Meaning Shift}} \\
\textbf{Explanation for} {\color{red3}\textbf{[3]}}: ``Airlines have forfeited self-regulation'' is presented as postulation in the text - the summary makes it seems like a fact. {\color{blue1}{Stating opinion as fact}}\\

\bottomrule
\end{tabular}
\caption{A dialogue example with generated summaries and human annotations in \textsc{TofuEval} (part 2). We show all three generated topics for the dialogue in the table. For each topic, we show one of the five generated summaries with human-written explanations and {\color{blue1}{error types}} for {\color{red3}{factually inconsistent sentences}}.} \label{tab:example-2}
\end{table*}

\label{sec:appendix}

\end{document}